\documentclass[runningheads]{llncs}
\usepackage[T1]{fontenc}
\usepackage[pdftex]{graphicx}
\usepackage{booktabs}
\usepackage[misc]{ifsym}

\usepackage{makecell}
\usepackage{subcaption}
\usepackage{mwe}

\usepackage[capitalise]{cleveref}
\crefname{subsection}{Subsection}{Subsections}
\usepackage{multirow}
\usepackage{algorithmic}
\usepackage{algorithm}

\begin{document}

\title{Policies of Multiple Skill Levels for Better Strength Estimation in Games}

\titlerunning{Policies of Multiple Skill Levels for Better Strength Estimation in Games}

\author{Kyota Kuboki \inst{1} \and
Tatsuyoshi Ogawa\inst{1} \and
Chu-Hsuan Hsueh\inst{1} \and
Shi-Jim Yen\inst{2} \and
Kokolo Ikeda\inst{1}}


\institute{Japan Advanced Institute of Science and Technology, Japan \email{\{s2460002,ogawa.tatsuyoshi,hsuech,kokolo\}@jaist.ac.jp}
\and
National Dong Hwa University, Taiwan \email{\{sjyen@gms.ndhu.edu.tw\}}
}

\maketitle              

\begin{abstract}
Accurately estimating human skill levels is crucial for designing effective human-AI interactions so that AI can provide appropriate challenges or guidance. In games where AI players have beaten top human professionals, strength estimation plays a key role in adapting AI behavior to match human skill levels. In a previous state-of-the-art study, researchers have proposed a strength estimator trained using human players' match data. Given some matches, the strength estimator computes strength scores and uses them to estimate player ranks (skill levels). In this paper, we focus on the observation that human players' behavior tendency varies according to their strength and aim to improve the accuracy of strength estimation by taking this into account. Specifically, in addition to strength scores, we obtain policies for different skill levels from neural networks trained using human players' match data. We then combine features based on these policies with the strength scores to estimate strength. We conducted experiments on Go and chess. For Go, our method achieved an accuracy of 80\% in strength estimation when given 10 matches, which increased to 92\% when given 20 matches. In comparison, the previous state-of-the-art method had an accuracy of 71\% with 10 matches and 84\% with 20 matches, demonstrating improvements of 8-9\%. We observed similar improvements in chess. These results contribute to developing a more accurate strength estimation method and to improving human-AI interaction.


\end{abstract}

\section{Introduction}
In recent years, artificial intelligence (AI) technology has been actively researched in various fields and has achieved remarkable results.
Especially in the field of games, AI players have surpassed the strength of top human professionals.
Examples of this include AlphaGo for the game of Go~\cite{Silver2016AlphaGo}, GT Sophy for the racing game Gran Turismo~\cite{Wurman2022GTSophy}, and AlphaStar for StarCraft II~\cite{Vinyals2019AlphaStar}.
Beyond research focused on enhancing strength, there is a growing interest in utilizing these powerful AI players for teaching or entertaining human players~\cite{Fujita2022AlphaDDA,Hsueh2024ApInGo,Liu2020Strength}.
One of the critical components for successfully entertaining and teaching human players is the ability to assess their skill levels.
Estimating a player's strength can be used for various purposes, such as preparing a suitable opponent AI, providing feedback to improve the player's skills, and designing game content with appropriate difficulty levels.

One of the most well-known methods for estimating player strength is the Elo rating system~\cite{Elo2008Rating}, which has been widely used in various competitive games.
However, since Elo ratings are calculated based only on match outcomes, a sufficient number of matches is necessary to achieve an accurate estimation.
To address this limitation, some techniques incorporate game-specific features.
For example, in chess, Tijhuis et al.~\cite{Tijhuis2023Predicting} used features such as the number of moves before casting, while in Go, Moud{\v{r}}{\'i}k et al.~\cite{Moudrik2015Evaluating} used features such as the number of stones captured by a move.
Other approaches aim for more general applicability by estimating player strength based on the degree of \emph{mistakes} in their moves, rather than relying solely on match outcomes~\cite{Guid2006Computer,Kosaka2018Examination}.
In these approaches, mistakes are evaluated using powerful AI players as a reference.

More recently, Chen et al.~\cite{Chen2025Strength} proposed a strength estimator based on the Bradley-Terry model.
This estimator determines a strength score for a given game state and move.
The strength scores from multiple moves in one or more games can be aggregated to estimate player strength.
In their experiments, they achieved over 80\% accuracy in predicting player ranks within 15 matches in Go and 26 matches in chess, setting a state-of-the-art result.

While the strength estimator achieved remarkable results, we identified areas for improvement.
Specifically, a player's overall skill level is usually influenced by multiple factors that may not be adequately captured by a single strength score.
For example, in Go and chess, skills such as positional judgment (e.g., maintaining good shape in Go or controlling the center in chess) and tactical calculation (e.g., accurately reading sequences of moves) may be evaluated differently.
Therefore, we consider it beneficial to consult with multiple models.
In this paper, we propose to estimate strength by utilizing (a) strength scores, as well as (b) policies (probability distributions for selecting moves) across different skill levels and (c) average score losses, which indicate the degree of mistakes made by players when compared to strong AI players.
We employ supervised learning to train models that take (a)--(c) as inputs and predict player ranks.
Our method achieved an accuracy of 80\% with 10 matches and 92\% with 20 matches in Go, with similar results observed in chess.
These results surpass the previous state-of-the-art method~\cite{Chen2025Strength}.


\section{Background}
\label{sec:background}

\subsection{Strength Estimation}

Elo rating system~\cite{Elo2008Rating} is a classical approach to estimating players' relative skill levels in competitive games.
Players' ratings are updated after each match based on the outcome--whether they won, drew, or lost.
Therefore, to achieve an accurate rating, a sufficient number of matches is necessary.
To address this limitation, various approaches have been proposed~\cite{Chen2025Strength,Guid2006Computer,Kosaka2018Examination,Liu2020Strength,Moudrik2015Evaluating,Moudrik2016Determining,Omori2024Chess,Tijhuis2023Predicting}.
One such approach involves using game-specific features that correlate with player skill.
For example, in chess, the timing of castling~\cite{Tijhuis2023Predicting} and, in Go, the number of stones captured per move~\cite{Moudrik2015Evaluating} have been examined.
Another approach evaluates a player's move quality by comparing their decisions with those of a strong AI player.
For example, the \emph{average loss} in scores resulting from the player's moves has been analyzed in chess~\cite{Guid2006Computer}, Go~\cite{Kosaka2018Examination}, and backgammon~\cite{UrlXGDocument}.
Additionally, some approaches employ deep neural networks that take game states, moves, etc., as inputs to predict player skill levels~\cite{Moudrik2016Determining,Omori2024Chess}.

Recently, Chen et al.~\cite{Chen2025Strength} proposed a strength estimator trained using human players' game records.
This estimator takes a game state and a move as inputs and outputs a strength score, the higher the stronger.
Assume that there are $R$ ranks, with $r_1$ being the strongest.
For a rank $r_i$, its strength score $\bar{\beta_i}$ is averaged from those of state-move pairs sampled from the rank.
The strength estimator was trained based on the Bradley-Terry model so that the average strength scores of the ranks are in descending order, i.e., $\bar{\beta_1}>\bar{\beta_2}>\ldots>\bar{\beta_R}$.
In addition, they introduced $r_\infty$, defined as the weakest rank, into the training process, aiming for a more robust estimation.
Their experiments showed that this method could predict player ranks with over 80\% accuracy in just 15 matches in Go and 26 matches in chess, setting a state-of-the-art result.

\subsection{Imitation of Human Moves Across Skill Levels}

Imitating human players' moves across various skill levels is crucial for developing AI players that can both entertain and teach humans. 
One effective approach to creating AI players that play human-like moves is supervised learning from human game records. 
For example, Maia~\cite{maia2020} consists of a set of models in chess that predict probability distributions of moves across different skill levels.
Each model was independently trained on game records corresponding to a specific player rating range. 
However, training multiple independent models can lead to inconsistencies among skill levels.
To address this, Maia-2~\cite{Tang2024Maia2} introduced a skill-aware attention mechanism, encoding player skill levels within a single unified model.
A similar approach was taken for Go, where KataGo HumanSL~\cite{UrlKataGoV15} used a single model to predict moves across different skill levels.
The inputs for this model include a game state, both players' ranks, and the match date.
Since these models output probability distributions over moves, they can be used not only to select the most likely move for a given skill level but also to evaluate how likely or unlikely a move is for players at that skill level.

\subsection{Ensemble Learning}

Ensemble learning techniques combine multiple models to obtain better prediction accuracy.
Stacking is one such technique that integrates base models by training a meta-model to make the final prediction based on the base models' outputs~\cite{Wolpert1992Stacked}.
For example, one may use decision trees to create some base models while using neural networks to create some more base models, followed by training a meta-model using linear regression to integrate these base models.
The stacking technique has been successfully applied to various fields, including medicine and spam filtering.
Moud{\v{r}}{\'i}k and Neruda~\cite{Moudrik2015Evolving} also explored stacking ensembles to estimate player strength and playstyle in Go, where the models took game-specific features~\cite{Moudrik2015Evaluating} as inputs.
By leveraging the complementary advantages of different approaches, meta-models created by the stacking technique usually yield better accuracy and robustness than individual base models.

\section{Method}
\label{sec:proposed method}

\begin{figure}[tb]
    \centering
    \includegraphics[width=0.98\linewidth]{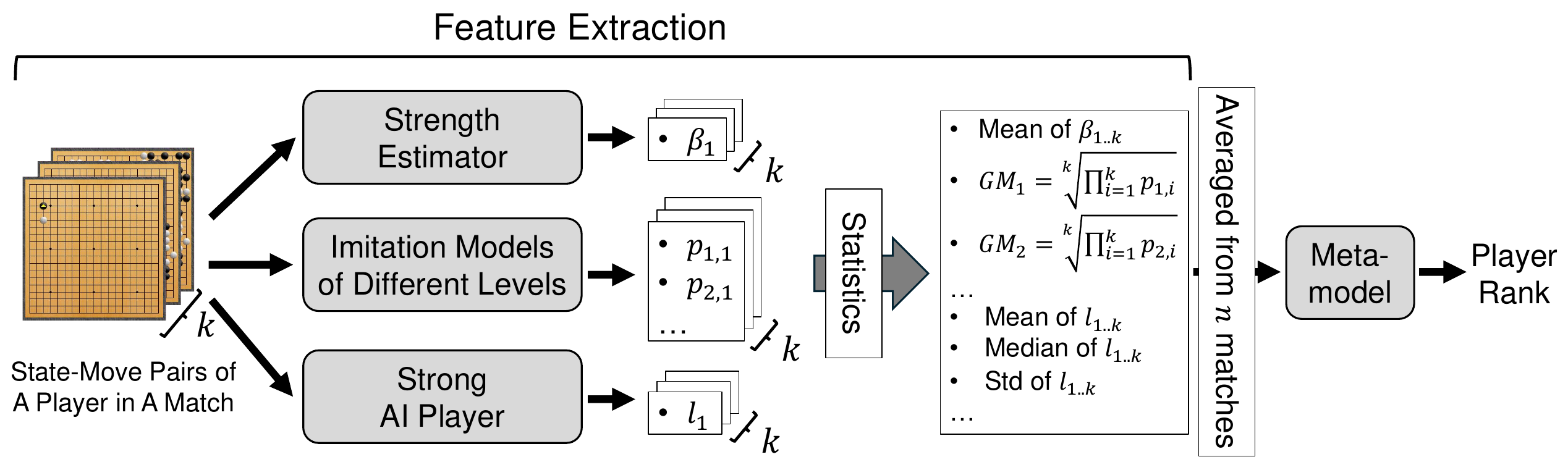}
    \caption{An overview of our method, where $k$ is the number of moves played by a player in a match, $\beta_i$ denotes strength score of state-move pair $(s_i, m_i)$, $p_{j,i}$ denotes the selection probability of move $m_i$ at state $s_i$ from the imitation model at skill level $j$, and $l_i$ denotes the loss of move $m_i$, defined as the difference between the state evaluations before and after executing $m_i$: $l_i = v'_i - v_i$. Here, $v_i$ is the evaluation of state $s_i$, and $v'_i$ is the evaluation of the resulting state after playing move $m_i$, both computed by a strong AI player.}
    \label{fig:overview}
\end{figure}

To achieve a more accurate strength estimation, we propose to create meta-models that integrate various models, including those that were not originally designed for strength estimation.
\cref{fig:overview} shows an overview of our method, including feature extraction from multiple models and rank estimation using these features.
In more detail, we utilize (a) strength scores predicted by Chen et al.'s strength estimator~\cite{Chen2025Strength}, (b) imitation models that output policies across different skill levels~\cite{UrlKataGoV15,maia2020}, and (c) game state evaluations based on strong AI players~\cite{UrlLc0,Wu2020KataGo}.
Among the three components (a) to (c), only (a) was originally designed for strength estimation.

Given a match, we separate the state-move pairs by the players.
For a player's state-move pairs $\{(s_i, m_i) | i\in[1, k]\}$ in a match, we collect features as follows, where $k$ is the number of moves the player made in the match.
First, we input each state-move pair to obtain the corresponding strength score $\beta_i$.
We then calculate the arithmetic mean of these strength scores using the formula $(\Sigma_{i=1..k}\beta_i)/k$.
This mean strength score is included as one of the features.

Next, assume that there are $L$ imitation models $\pi_1$, $\pi_2$, \ldots, $\pi_L$ corresponding to skill levels $1$ to $L$.
Note that the $L$ skill levels do not need to be the same as the target rank to predict.
For example, in our experiments in chess, our target ranks to predict cover ratings from R1000 to R2599, while the imitation models we employ cover ratings from R1100 to R1999.
For each skill level $j$, we input each state $s_i$ to obtain the move probability distribution at skill level $j$ and extract the selection probability (or prior) of the played move $m_i$, denoted as $p_{j, i}=\pi_j(m_i | s_i)$.
We then calculate the geometric mean of these priors, i.e., likelihood, at skill level $j$ using the formula ${GM}_j=\sqrt[k]{\prod_{i=1}^{k} p_{j, i}}$.
Namely, the features include $L$ geometric means of the priors for different skill levels.

The main motivation for using imitation models across different skill levels will be explained in more detail with evidence in \cref{subsec:discussions main}.
Briefly speaking, when distinguishing strong players within a narrow skill range (e.g., 9 dan and 8 dan players in Go), selection probabilities from models at similarly high skill levels (e.g., 7 dan) often fail to provide sufficient separation.
In contrast, models at relatively low skill levels (e.g., 1 dan) can clearly distinguish those strong players by how unlikely the moves are made.
Therefore, employing priors from imitation models across different skill levels ultimately help improve the accuracy of strength estimation.

Finally, assume that there is a strong AI player that can provide accurate state evaluations.
For each move-state pair $(s_i, m_i)$, we calculate the loss $l_i$ as $v'_i - v_i$, where $v_i$ is the evaluation of $s_i$, and $v'_i$ is the evaluation of the resulting state after playing $m_i$.
We then obtain the arithmetic mean of the losses, denoted by $MeanLoss$, along with the median $MedLoss$ and standard deviation $StdLoss$, all computed using only the moves within the first $N_{cut}$ plies in the match.
We include losses in the features because we consider that losses can reflect players' ability in reading.
In \cref{subsec:discussions main}, more discussions will be made on losses, especially regarding $N_{cut}$.

Given a set of match data with the players' ranks being labeled, we first collect the features described earlier.
These features, along with players' ranks, are used to train meta-models based on supervised learning for rank estimation.

\section{Common Settings in Our Experiments}
\label{sec:common setting}

In this section, we describe the common settings for the following experiments in Go and chess.

\subsection{Match Data}
\label{subsec:match data}

For our experiments, we collected game records of Go from an online Go server called Fox Go~\cite{FoxGo} and game records of chess from an online chess server called Lichess~\cite{Lichess}.
The selection criteria were similar to those of Chen et al.~\cite{Chen2025Strength}, with detailed information provided in \cref{appendix:selection criteria}.
In addition, we followed their rank grouping.
For Go, there were 11 rank groups, from the weakest to the strongest being 3--5 kyu (3--5k), 1--2 kyu (1--2k), 1 dan (1d), 2 dan (2d), \ldots, 8 dan (8d), and 9 dan (9d).
For chess, there were 8 rank groups separated according to player ratings.
The rank groups from the weakest to the strongest were R1000--R1199, R1200--R1399, \ldots, R2200--R2399, and R2400--R2599.\footnote{Chen et al.~\cite{Chen2025Strength} numbered $R$ ranks $r_1, r_2, \ldots, r_R$ from strong to weak.
In this paper, we number rank groups $g_0, g_1, \ldots, g_{R-1}$ from weak to strong.}

For Go, we collected 5,000 matches for each of the 11 rank groups, using 55,000 matches in total.
For chess, we collected 13,000 matches for each of the 8 rank groups, using 104,000 matches in total.
From each match, we further separated black and white players into different data points, resulting in a total of 110,000 data points for Go and 208,000 data points for chess.

\subsection{Feature Extraction}

As explained in \cref{sec:proposed method}, we utilize strength estimators, imitation models of different skill levels, and strong AI players to estimate strength.
In this subsection, we specify the models used in our experiments.

For strength estimators, we employed the models of Go and chess released by Chen et al.~\cite{Chen2025Strength}, specifically, those with an additional rank $r_\infty$ used in the training process.
For imitation models, we employed KataGo HumanSL~\cite{UrlKataGoV15} for Go and Maia~\cite{maia2020} for chess.
In more detail, in Go, we obtained policies representing 10 skill levels from KataGo HumanSL by setting the ranks to 10k, 8k, \ldots, 2k, 1d, 3d, \ldots, 9d, which fully covers the prediction targets 5k, \ldots, and 9d.
In chess, we obtained policies representing 9 skill levels from all the released models, Maia-1100, -1200, \ldots, -1800, and -1900, which does not fully cover the prediction targets R1000--R2599.
As explained in \cref{sec:proposed method}, the skill levels of the imitation models do not need to match the rank groups to predict, though it is preferable that the rank groups are fully covered by the skill levels.

For strong AI players, we employed neural networks trained based on AlphaZero from KataGo~\cite{Wu2020KataGo} for Go and Leela Chess Zero~\cite{UrlLc0} for chess.
More specifically, for state evaluations in Go, we used \texttt{scoreLead} predicted by the network, which estimates the number of points the current player is leading in territory.
For chess, we first obtained the win rate ${wr}_i$ predicted by the network for a given state $s_i$.
We then applied a logit transformation to derive the state evaluation $v_i$ by
\begin{equation}
\label{eq: winrate logit}
    v_i=\log(\frac{{wr}_i}{1-{wr}_i}).
\end{equation}
The reason for applying the logit transformation is explained as follows.
When the winners and losers are clear-cut, the fluctuations in win rates tend to be small.
Even if a player makes a terrible mistake that does not alter the match outcome, the changes in win rates remain small and fail to adequately reflect the impact of that mistake.
By using the logit transformation, we can better capture the mistakes that are made in the endgame phase.

After obtaining losses $l_i$ for all moves $m_i$, we extracted loss-related features $MeanLoss$, $MedLoss$, and $StdLoss$ computed using the moves within the first $N_{cut}$ plies, as discussed in \cref{sec:proposed method}.
We considered $N_{cut}\in\{50, 100, \infty\}$, where $N_{cut}=\infty$ indicates that all moves by the player in the match are included.

\subsection{Training of Rank Estimation Meta-models}

Assume that there are $R$ rank groups, from the weakest to strongest being $g_0, g_1, \ldots, g_{R-1}$.
From each group $g_j$, we collected features from the matches, along with $j$ as the rank label, to train regression models.
In more detail, we repeat the following process 10,000 times in Go and 26,000 times in chess: Sample $n$ data points, average the features, and add the pair (the average features, $j$) to training data.
In the experiments, we considered $n \in \{1,5,10,15,20 \}$.
We employed LightGBM~\cite{UrlLightGBM} with default settings for regression model training.


\section{Main Results}
\label{sec:main results}

In this section, we present the main results of our rank estimation models and compare them to Chen et al.'s method~\cite{Chen2025Strength}.

\subsection{Testing Data and Evaluation Procedure}
\label{subsec:testing main}

For testing data, we collected game records of Go and chess using the same selection criteria described in \cref{subsec:match data}.
For Go, we collected 900 matches for each of the 11 rank groups, using 9,900 matches in total. 
From each match, we randomly sampled a player and added that player's data point to the testing data, leading to a total of 9,900 data points.
We applied a similar process to chess, with 1,200 matches for each of the 8 rank groups, and obtained 9,600 data points.
The matches collected for training and testing were distinct sets.

Both Chen et al.'s method~\cite{Chen2025Strength} and our method were evaluated through a consistent testing procedure following Chen et al.'s work.
For each rank group $g_j$, both methods predicted ranks based on $n$ randomly sampled data points, which was repeated 500 times to ensure a stable evaluation.
In other words, with $R$ rank groups, each method made predictions $500R$ times.
In the experiments, we considered $n\in \{1,5,10,15,20 \}$.

For our method, the features of the $n$ sampled games were averaged and then inputted to the rank estimation meta-models that were trained with the corresponding $n$.
Since the meta-models are regression models, we rounded the predicted real values to the nearest integers, which then served as the predicted rank groups.
In addition, we performed preliminary experiments to select features related to loss.
As a result, we used $MeanLoss$ with $N_{cut}=100$ and $MedLoss$  with $N_{cut}=\infty$ for Go and $MeanLoss$ and $StdLoss$ with $N_{cut}=50$ for chess.
We did not use all combinations of $MeanLoss$, $MedLoss$, $StdLoss$, and $N_{cut} \in \{50, 100, \infty\}$ because overfitting might occur due to using too many highly correlated features.

Accuracy was used as a straightforward evaluation metric for the prediction results, which was defined as the ratio that the predicted rank groups matched the actual rank groups.
Additionally, we considered the metric of \emph{accuracy$\pm1$} as Chen et al.~\cite{Chen2025Strength} did, which treats predictions of $g_{j-1}$ and $g_{j+1}$ as accurate if the actual rank group is $g_j$.
The accuracy$\pm1$ metric is reasonable, considering that human players' performances may vary.

\subsection{Rank Estimation Accuracy}
\label{subsec:accuracy main}

\cref{tab:result1} shows the rank estimation accuracy of our method compared to that of Chen et al.~\cite{Chen2025Strength} in both Go and chess.
The confusion matrices are shown in \cref{appendix:confusion matrix group}.
With 20 data points, our method achieved an accuracy of 91.7\% in Go and 86.2\% in chess, surpassing Chen et al.'s method by 7.9\% and 6.7\%, respectively.
Additionally, for accuracy$\pm1$, our method achieved 82.8\% in Go and 78.8\% in chess with only one data point.
Even with a single data point, our method could effectively estimate a player's rank.

{
\tabcolsep = 10pt

\begin{table}[bt]
    \centering
    \caption{Rank estimation accuracy using $n$ matches (accuracy$\pm1$ in parentheses)}
    \label{tab:result1}
    \begin{tabular}{c|cc|cc}
        \toprule
        \multirow{2}{*}{$n$} & \multicolumn{2}{c|}{Go} & \multicolumn{2}{c}{Chess}\\
        \cmidrule(ll){2-5}
        & Proposed & Chen et al.~\cite{Chen2025Strength} & Proposed & Chen et al.~\cite{Chen2025Strength}\\
        \midrule
        1 & \textbf{37.5} (\textbf{82.8}) & 33.7 (74.0) & \textbf{32.2} (\textbf{78.8}) & 31.8 (70.5)\\
        5 & \textbf{64.8} (\textbf{98.8}) & 56.3 (96.6) & \textbf{58.7} (\textbf{97.2}) & 51.7 (94.9)\\
        10 & \textbf{79.6} (\textbf{99.9}) & 71.4 (99.7) & \textbf{74.3} (\textbf{99.7}) & 66.5 (99.1)\\
        15 & \textbf{87.5} (\textbf{100}) & 78.2 (99.9) & \textbf{81.5} (\textbf{100}) & 73.6 (99.7)\\
        20 & \textbf{91.7} (\textbf{100}) & 83.8 (\textbf{100}) & \textbf{86.2} (\textbf{100}) & 79.5 (99.9)\\
        \bottomrule
    \end{tabular}
\end{table}
}

{
\tabcolsep = 10pt

\begin{table}[bt]
    \centering
    \caption{Ablation study of our rank estimation meta-model in Go}
    \label{tab:ablation1_go}
    \begin{tabular}{c|cccc}
        \toprule
        $n$ & Use All & w/o Strength & w/o Prior & w/o Loss\\
        \midrule
        1 & 37.5 (\textbf{82.8}) & 34.4 (77.8) & 32.3 (77.3) & \textbf{37.9} (82.4)\\
        5 & \textbf{64.8} (\textbf{98.8}) & 60.0 (96.9) & 58.1 (97.7) & \textbf{64.8} (\textbf{98.8})\\
        10 & 79.6 (\textbf{99.9}) & 73.5 (99.5) & 73.0 (99.7) & \textbf{79.8} (\textbf{99.9})\\
        15 & \textbf{87.5} (\textbf{100}) & 82.0 (99.8) & 81.1 (99.9) & 87.4 (\textbf{100})\\
        20 & 91.7 (\textbf{100}) & 87.7 (\textbf{100}) & 86.4 (\textbf{100}) & \textbf{91.9} (\textbf{100})\\
        \bottomrule
    \end{tabular}
\end{table}
}
{
\tabcolsep = 10pt

\begin{table}[bt]
    \centering
    \caption{Ablation study of our rank estimation meta-model in chess}
    \label{tab:ablation1_chess}
    \begin{tabular}{c|cccc}
        \toprule
        $n$ & Use All & w/o Strength & w/o Prior & w/o Loss\\
        \midrule
        1 & \textbf{32.2} (\textbf{78.8}) & 27.5 (72.8) & 28.3 (75.7) & 31.5 (78.2)\\
        5 & \textbf{58.7} (97.2) & 51.0 (94.7) & 54.0 (96.3) & 58.6 (\textbf{97.3})\\
        10 & \textbf{74.3} (99.7) & 63.7 (98.9) & 69.5 (99.5) & 72.9 (\textbf{99.8})\\
        15 & \textbf{81.5} (\textbf{100}) & 72.8 (99.7) & 77.3 (99.9) & 81.1 (\textbf{100})\\
        20 & \textbf{86.2} (\textbf{100}) & 78.5 (99.8) & 82.2 (\textbf{100}) & 85.0 (\textbf{100})\\
        \bottomrule
    \end{tabular}
\end{table}
}

Furthermore, we conducted ablation studies to investigate how accuracy was affected by each set of features from strength scores, priors from imitation models, and losses.
The results are shown in \cref{tab:ablation1_go,tab:ablation1_chess}.
When strength scores or priors were excluded, the accuracy dropped drastically in both Go and chess.
In contrast, removing the losses had a minor effect, with accuracy in Go remaining almost the same and only a slight decrease observed in chess.
The results suggested that strength scores and priors contributed significantly to accurate rank estimation.
Interestingly, when only using priors and losses (i.e., the ``w/o Strength'' column in \cref{tab:ablation1_go,tab:ablation1_chess}), the accuracy in Go was already higher than Chen et al.'s method~\cite{Chen2025Strength}, while achieving similar accuracy in chess.

\subsection{Discussions}
\label{subsec:discussions main}

In this subsection, we first investigate why prior geometric means across different skill levels contributed to rank estimation, followed by a discussion on potential issues related to losses.

\subsubsection{Effects of Multiple Priors Across Skill Levels}

\begin{figure}[bt]
    \centering
    \begin{subfigure}[t]{0.49\linewidth}
        \centering
        \raisebox{3mm}{\includegraphics[width=59mm]{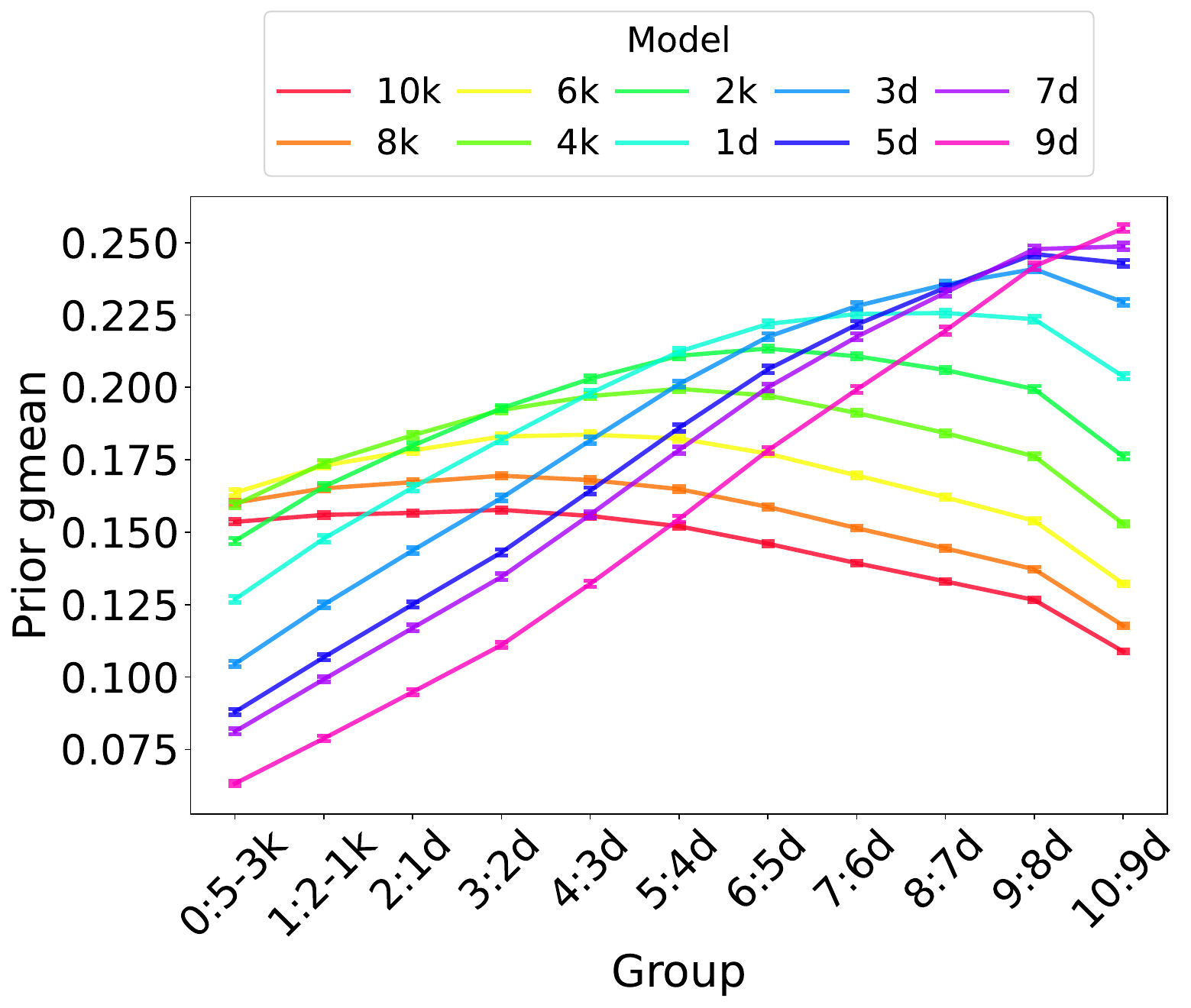}}
        \caption{Go}
        \label{fig:plot_prior_gmean_go}
    \end{subfigure}
    \begin{subfigure}[t]{0.49\linewidth}
        \centering
        \includegraphics[width=59mm]{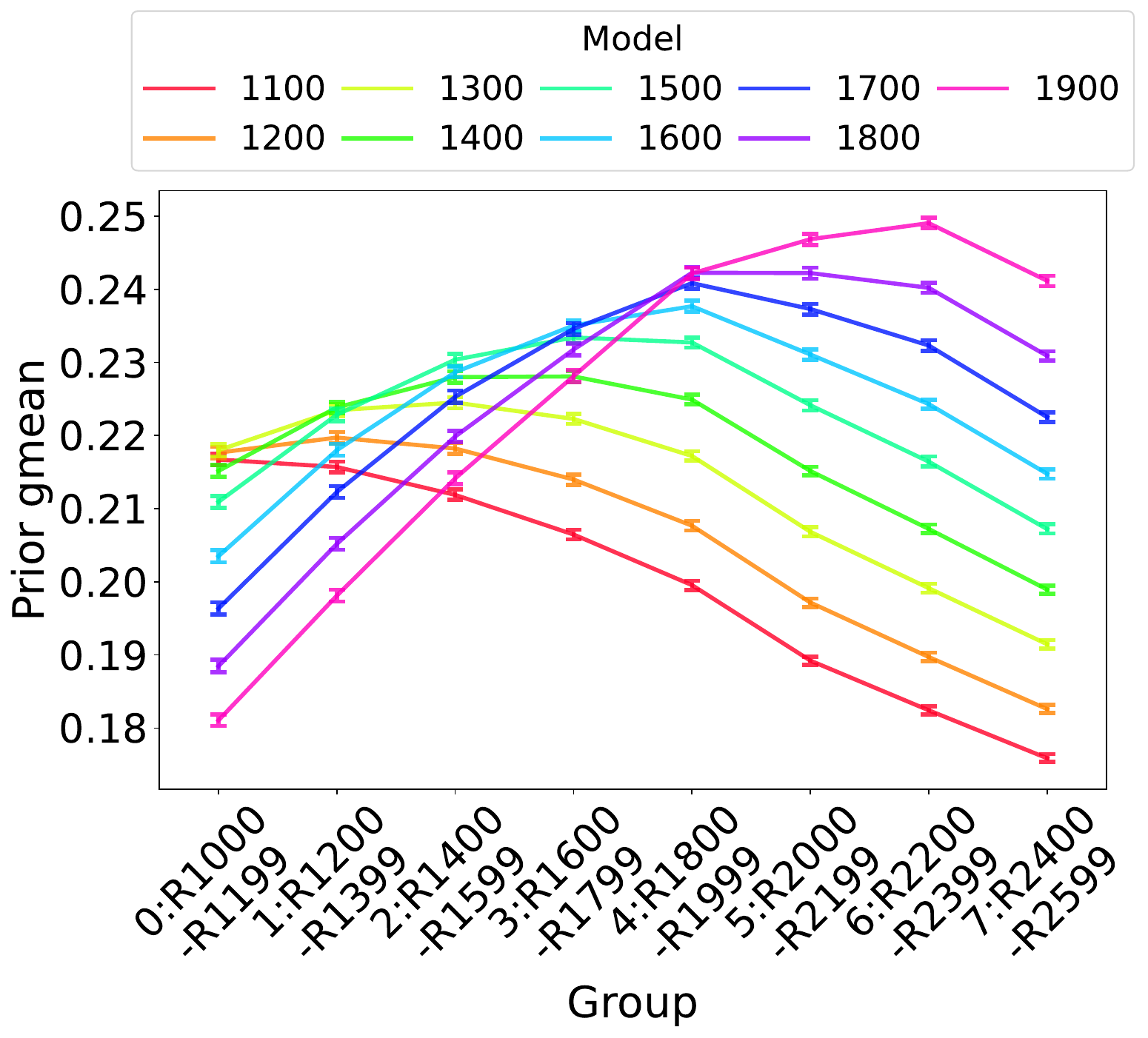}
        \caption{Chess}
        \label{fig:plot_prior_gmean_chess}
    \end{subfigure}
    \caption{The geometric means of priors, along with 95\% confidence intervals, across various skill levels for different rank groups.}
    \label{fig:plot_prior_gmean}
\end{figure}

\cref{fig:plot_prior_gmean} illustrates the geometric means of priors across various skill levels in Go and chess for each rank group.
Taking \cref{fig:plot_prior_gmean_go} as an example, the x-axis represents rank groups from 0 (3--5k, weakest) to 10 (9d, strongest).
The y-axis represents the geometric mean of priors, derived from the state-move pairs in the training data described in \cref{subsec:match data}.
Each curve corresponds to a specific skill level, such as 10k, 8k, and up to 9d, from which the priors were obtained.

When focusing on a single curve, we can calculate the slopes for each pair of the adjacent rank groups.
A steeper slope suggests that the imitation model at that skill level is more effective in distinguishing between the two rank groups.
For example, assume that we want to distinguish 8d and 9d players in Go.
We would expect the 2k or 1d model to be more informative than the 5d or 7d model.
Specifically, for both 5d and 7d models, the prior geometric means for the 8d and 9d rank groups were very close.
In contrast, for both 2k and 1d models, the prior geometric means for the 9d group was significantly lower than that of the 8d group.
This indicates that the moves played by 9d players were even less likely to be seen from 2k or 1d players than those played by 8d players.
From another viewpoint, this time, we take chess as an example.
The Maia-1900 model might be useful to distinguish players within rank groups 0 to 3 (R1000 to R1799), while being less useful to distinguish players within rank groups 4 to 7 (R1800 to R2599).
On the contrary, the Maia-1500 model might be useful for distinguishing rank groups 4 to 7 but not for 0 to 3.
Experiments presented in \cref{appendix:maia1500 maia1900} confirmed this hypothesis.

In addition to the observation that different imitation models are useful to distinguish different rank groups, we found another key benefit.
The priors associated with each player's moves fluctuate to some extent, as the priors are influenced by the given states and probably the player's thought at that time, causing the prior geometric means to vary even for the same player.
To reduce random fluctuations, it may be effective to use multiple indicators that are related to skill levels but not overly correlated with each other.
For example, in \cref{fig:plot_prior_gmean_chess}, Maia-1100 and Maia-1200 show similar tendencies, and one may expect that using priors from either model would be sufficient.
However, when using priors only from Maia-1100 for rank estimation with $n=20$, the accuracy was 28.1\%, and using only Maia-1200 resulted in 24.3\%.
In contrast, when using priors from both Maia-1100 and Maia-1200, the accuracy was improved to 35\%.

Even combining models with relatively strong positive correlations, like Maia-1100 and Maia-1200, yielded accuracy improvement.
Given this, it is reasonable to expect that utilizing more diverse models, such as Maia-1900 or strength estimators could further improve the estimation accuracy.

\subsubsection{Potential Issues Related to Losses}

\begin{figure}[bt]
    \centering
    \begin{subfigure}[t]{0.49\linewidth}
        \centering
        \includegraphics[height=40mm,width=55mm]{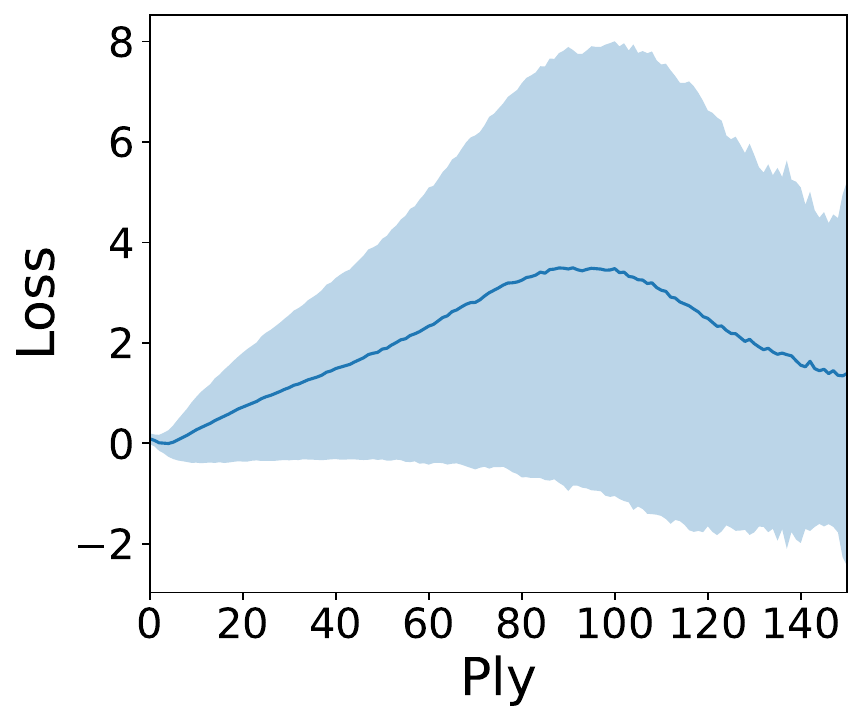}
        \caption{Go}
        \label{fig:plot_loss_mean_go} 
    \end{subfigure}
    \begin{subfigure}[t]{0.49\linewidth}
        \centering
        \includegraphics[height=40mm,width=55mm]{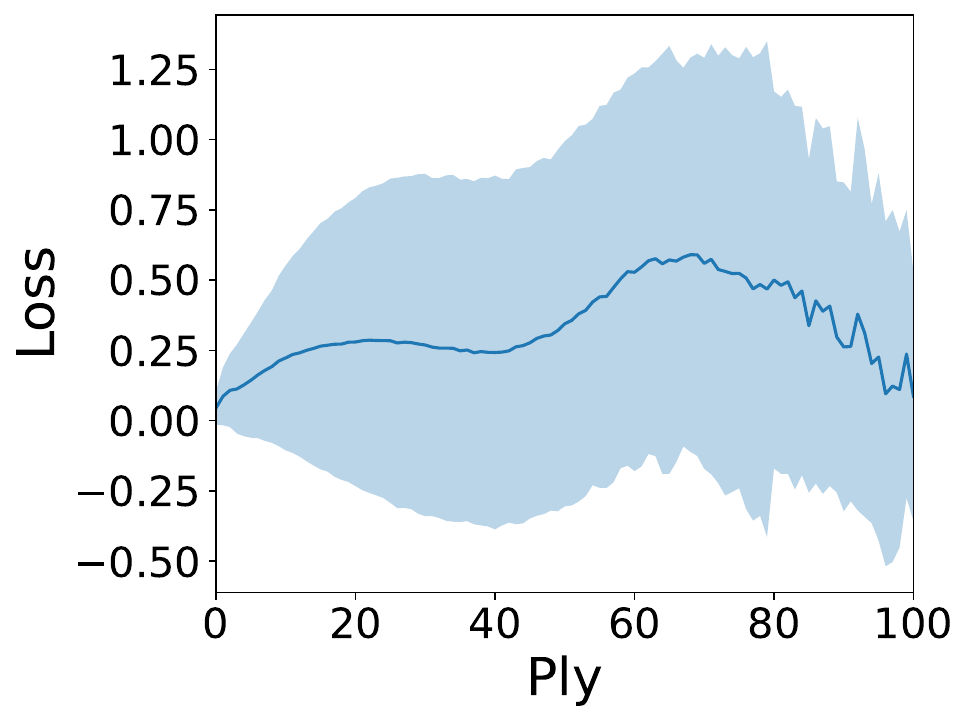}
        \caption{Chess}
        \label{fig:plot_loss_mean_chess} 
    \end{subfigure}
    \caption{The mean loss at each ply.}
    \label{fig:plot_loss_per_move}
\end{figure}

We included losses in the features because we considered that losses could reflect players' ability in reading, thereby contributing to rank estimation.
However, the results in \cref{subsec:accuracy main} showed that the impact of losses was quite limited.
We suspected that this might be due to the significant variation in losses across different matches.

\cref{fig:plot_loss_per_move} illustrates the mean loss at each ply for both Go and chess, based on the state-move pairs in the training data described in \cref{subsec:match data}.
The shaded areas represent the standard deviations.
In both Go and chess, the loss distributions exhibit mountain-shaped patterns.
In Go, the peak occurred around the 100th ply, while in chess, two peaks occurred around the 20th and 70th plies.
The standard deviations reached their maximum at these peaks in both games.

We interpreted the distributions as follows.
In the opening phase, the situation remained relatively stable across all matches.
However, in the middle phase, game states varied significantly depending on the match, likely due to increased complexity and strategic divergence.
In the endgame phase, the situation became stable again, likely because the outcomes got clearer.
As a result, simply averaging the losses over an entire match might lead to underestimation in matches that extended into the endgame phase.
This suggests that setting a threshold $N_{cut}$ to focus on a specific range from the opening phase was effective in improving player rank estimation.

\section{Rank Estimation of Individual Players}
\label{sec:estimation by player}

In the experiments presented in \cref{sec:main results}, we followed the evaluation procedure of Chen et al.~\cite{Chen2025Strength}, where rank estimation was performed by randomly sampling $n$ matches from a given rank group.
However, in practical applications, we are often interested in estimating a player's rank based solely on their own matches.

When switching from random sampling to player-specific sampling, two key differences arise.
First, even within the same rank group, individual players have unique playstyles and weaknesses.
For example, some players may be good at positional intuition but struggle with deep reading, leading to frequent mistakes, whereas others may have weaker positional judgment but compensate with strong reading ability.
As a result, the variance in feature distributions increases, making rank estimation more challenging.

Second, random sampling can lead to overestimation due to averaging effects.
For example, let's consider the R1600--R1799 rank group in chess and 20 matches collected from a single player rated R1610.
For the rank estimation model to correctly predict this rank, the acceptable error range of rating is $[-11, +189]$, which means the smallest allowable margin of error is 11 rating points.
In contrast, if we randomly sample 20 matches from the R1600--R1799 rank group, the average rating of these matches is likely to be around R1700.
In this case, even with a prediction error of $\pm$50 rating points, the model could still estimate the rank group correctly.
In other words, the acceptable estimation error differs significantly between the two evaluation procedures.
To eliminate these influences, we evaluated rank estimation models under player-specific sampling to assess the practical accuracy achievable in real-world scenarios.

\subsection{Testing Data and Evaluation Procedure}

For testing data, we collected game records of Go and chess using almost the same selection criteria described in \cref{subsec:match data}.
The only difference was that for rank estimation of individual players, we added a condition: The match must involve at least one player who has played a minimum of 20 matches in a rank group.
For each rank group, we collected matches from 100 players, each with 20 data points.
As a result, we had a total of 22,000 data points for Go across 11 rank groups and 16,000 for chess across 8 rank groups.

With these new sets of testing data, we conducted evaluations based on both random sampling and player-specific sampling to investigate the gap.
The evaluations based on random sampling followed the same procedure described in \cref{subsec:testing main}.
The evaluations based on player-specific sampling followed a similar procedure, where the only difference lies in the sampling.
For each player in a rank group, the models predicted ranks based on $n$ randomly sampled data points from that player, which was repeated 5 times.
In other words, with $R$ rank groups, each model made predictions $5\times 100 \times R$ times, which was the same as random sampling.
In the experiments, we considered $n \in \{5, 10, 15\}$.
The reason for excluding $n=1$ was that random sampling and player-specific sampling are identical with only one data point.
Additionally, the reason for excluding $n=20$ was that we only collected 20 data points for each player.
Repeating 5 times to sample 20 (i.e., all) data points for a player always produces the same result.

The rank estimation meta-models for our method used in this experiment were the same as those in \cref{sec:main results}.
In other words, we did not create new meta-models specifically for rank estimation of individual players.
While one might consider developing specific meta-models for this task, we believe that additional techniques are necessary to improve the performance, particularly as the size of the training data decreases.

\subsection{Rank Estimation Accuracy}

\cref{tab:result2_go,tab:result2_chess} show the rank estimation accuracy of our method compared to that of Chen et al.~\cite{Chen2025Strength} in Go and chess, respectively, under both scenarios of random sampling and player-specific sampling.
The confusion matrices are shown in \cref{appendix:confusion matrix player}.
Under the player-specific sampling scenario, our method achieved an accuracy of 63.2\% in Go and 60.4\% in chess with 15 data points.
Compared to Chen et al.'s method, the accuracy was 1.9\% and 9.7\% higher in Go and chess, respectively.
The improvement in accuracy for chess was especially significant.
The difference in the improvement between Go and chess will be discussed in \cref{subsec:discussions by player}.

{
\tabcolsep = 10pt
\begin{table}[bt]
    \centering
    \caption{Rank estimation accuracy using $n$ matches (accuracy$\pm1$ in parentheses) in Go based on the testing data in \cref{sec:estimation by player}}
    \label{tab:result2_go}
    \begin{tabular}{c|cc|cc}
    \toprule
    \multirow{2}{*}{$n$} & \multicolumn{2}{c|}{Random Sampling} & \multicolumn{2}{c}{Player-Specific Sampling} \\ \cmidrule(ll){2-5} 
     & Proposed & Chen et al.~\cite{Chen2025Strength} & Proposed & Chen et al.~\cite{Chen2025Strength}\\ \hline
    5 & \textbf{70.4} (\textbf{99.5}) & 59.8 (97.5) & \textbf{58.1} (\textbf{95.3}) & 51.6 (93.1)\\
    10 & \textbf{82.8} (\textbf{99.9}) & 74.1 (99.7) & \textbf{61.0} (\textbf{96.2}) & 59.4 (96.0)\\
    15 & \textbf{89.4} (\textbf{100}) & 81.3 (\textbf{100})& \textbf{63.2} (96.6) & 61.1 (\textbf{96.9})\\
    \bottomrule
    \end{tabular}
\end{table}
}

{
\tabcolsep = 10pt
\begin{table}[bt]
    \centering
    \caption{Rank estimation accuracy using $n$ matches (accuracy$\pm1$ in parentheses) in chess based on the testing data in \cref{sec:estimation by player}}
    \label{tab:result2_chess}
    \begin{tabular}{c|cc|cc}
    \toprule
    \multirow{2}{*}{$n$} & \multicolumn{2}{c|}{Random Sampling} & \multicolumn{2}{c}{Player-Specific Sampling} \\ \cmidrule(ll){2-5} 
     & Proposed & Chen et al.~\cite{Chen2025Strength} & Proposed & Chen et al.~\cite{Chen2025Strength}\\ \hline
    5 & \textbf{57.7} (\textbf{97.9}) & 51.6 (94.3) & \textbf{52.2} (\textbf{94.5}) & 44.4 (87.9)\\
    10 & \textbf{70.6} (\textbf{99.8}) & 61.7 (98.6) & \textbf{57.0} (\textbf{97.2}) & 50.0 (91.5)\\
    15 & \textbf{78.1} (\textbf{100}) & 67.7 (99.5)& \textbf{60.4} (\textbf{97.7}) & 50.8 (92.7)\\
    \bottomrule
    \end{tabular}
\end{table}
}

When comparing random sampling to player-specific sampling, we observed significant differences in accuracy, though the differences in accuracy$\pm1$ were relatively minor.
For example, with 15 data points, our method showed an accuracy difference of 26.2\% in Go and 17.7\% in chess between the two scenarios, which was 3.4\% and 2.3\% for accuracy$\pm1$.
To sum up, the results confirmed the gap between random sampling and player-specific sampling for rank estimation.

Similar to \cref{subsec:accuracy main}, we conducted ablation studies under the player-specific sampling scenario to investigate how accuracy was affected by each set of features from strength scores, priors from imitation models, and losses.
Detailed results are presented in \cref{appendix:ablation2}, where the tendency was similar to \cref{subsec:accuracy main}.
Briefly speaking, when strength scores or priors were excluded, the accuracy dropped drastically.
When excluding losses, the accuracy drops were still relatively minor but were bigger and more frequent compared to \cref{tab:ablation1_go,tab:ablation1_chess}.
We believe this was because losses contributed to evaluating players' reading ability, which became increasingly important for accurately estimating the ranks of individual players.


\subsection{Discussions}
\label{subsec:discussions by player}

\begin{figure}[bt]
    \centering
    \begin{subfigure}[t]{0.49\linewidth}
        \centering
        \raisebox{2mm}{\includegraphics[width=55mm]{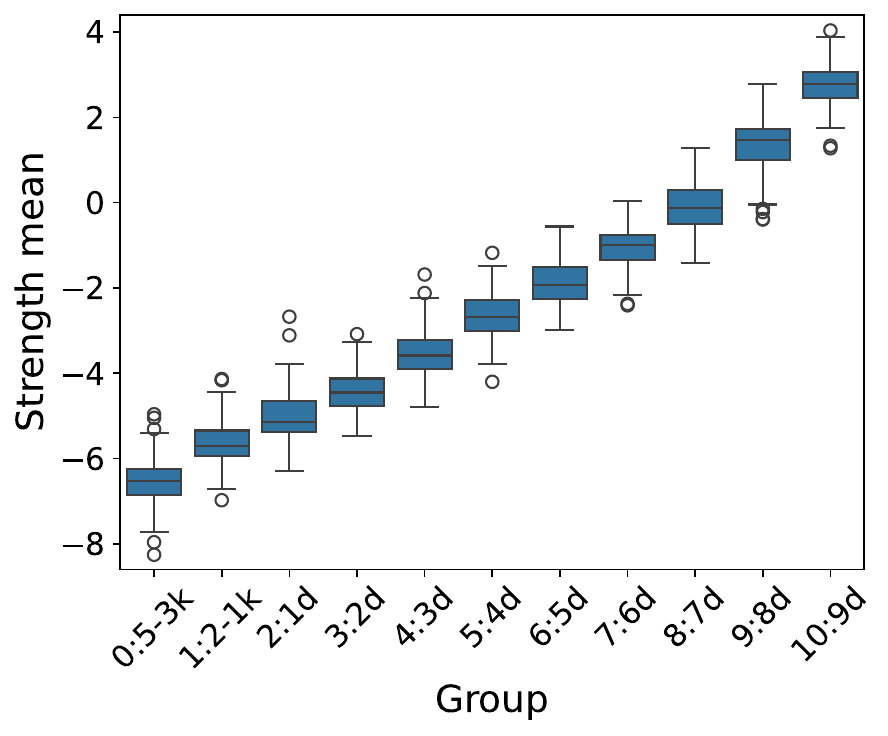}}
        \caption{Go}
        \label{fig:boxplot_str_gmean_go}
    \end{subfigure}
    \begin{subfigure}[t]{0.49\linewidth}
        \centering
        \includegraphics[width=55mm]{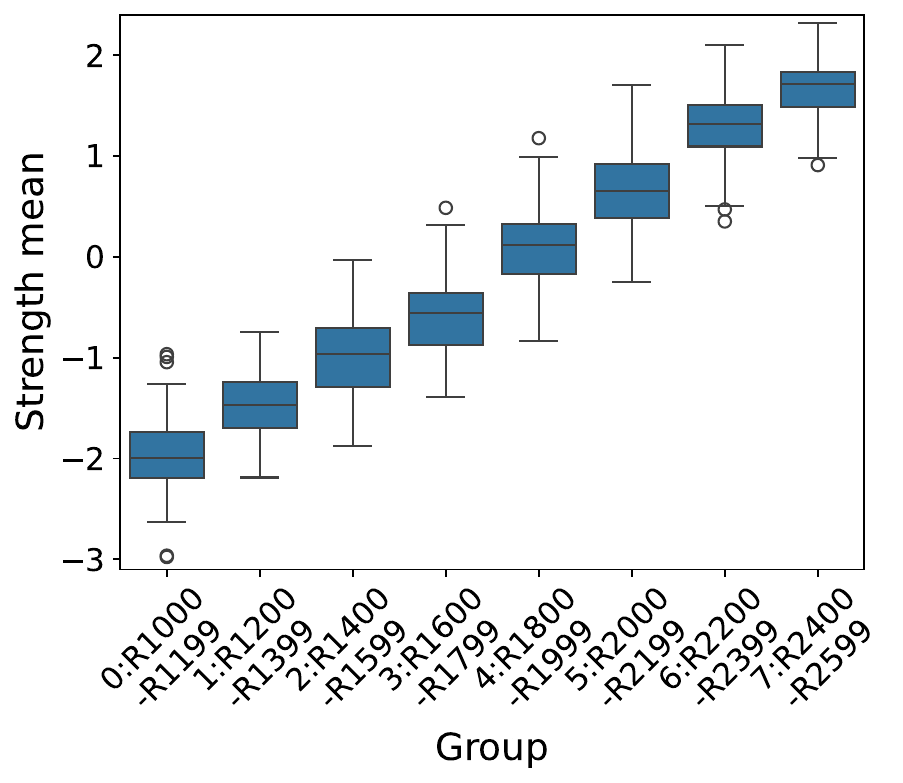}
        \caption{Chess}
        \label{fig:boxplot_str_gmean_chess}
    \end{subfigure}
    \caption{The mean strength scores of each player in each rank group with $n=20$.}
    \label{fig:boxplot_str_gmean}
\end{figure}

In this subsection, we examine the degree of variation in individual players' feature values within the same rank group.
\cref{fig:boxplot_str_gmean} shows box plots illustrating the distribution of mean strength scores for 20 sampled data points from each of the 100 players within every rank group.
In both Go and chess, rank groups of stronger players generally had higher mean strength scores.
However, chess (\cref{fig:boxplot_str_gmean_chess}) tended to show greater overlap between adjacent rank groups compared to Go (\cref{fig:boxplot_str_gmean_go}).
This explains the results in \cref{tab:result2_chess}, where using priors or losses in addition to strength scores led to much better rank estimation in chess.
Box plots of prior geometric means and losses are shown in \cref{appendix:boxplot}.

\begin{figure}[bt]
    \centering
    \begin{subfigure}[t]{0.49\linewidth}
        \centering
        \raisebox{2mm}{\includegraphics[width=55mm]{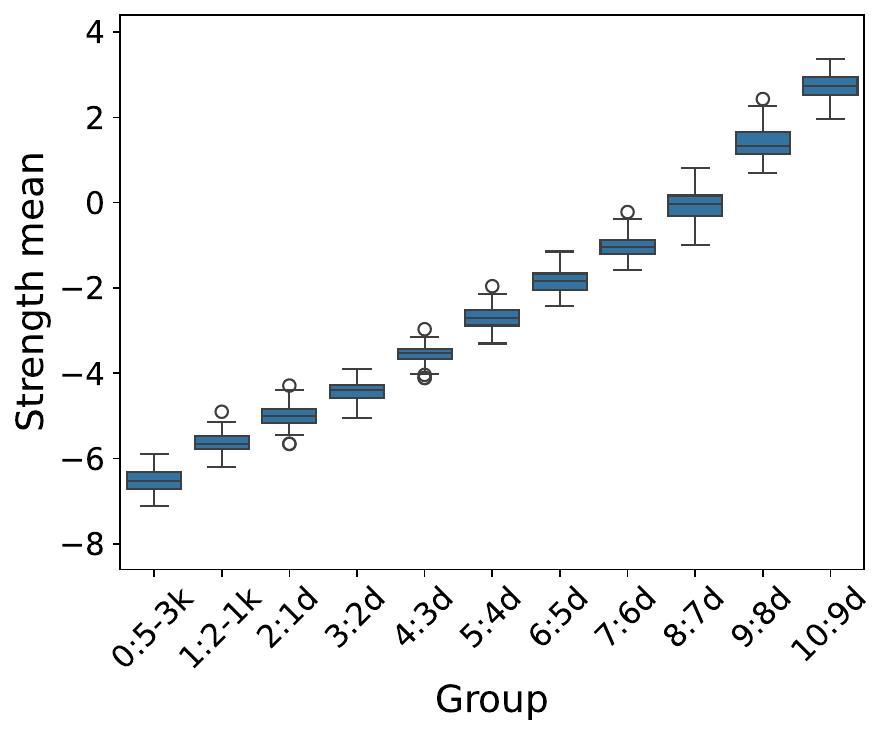}}
        \caption{Go}
        \label{fig:boxplot_str_gmean_go_random}
    \end{subfigure}
    \begin{subfigure}[t]{0.49\linewidth}
        \centering
        \includegraphics[width=55mm]{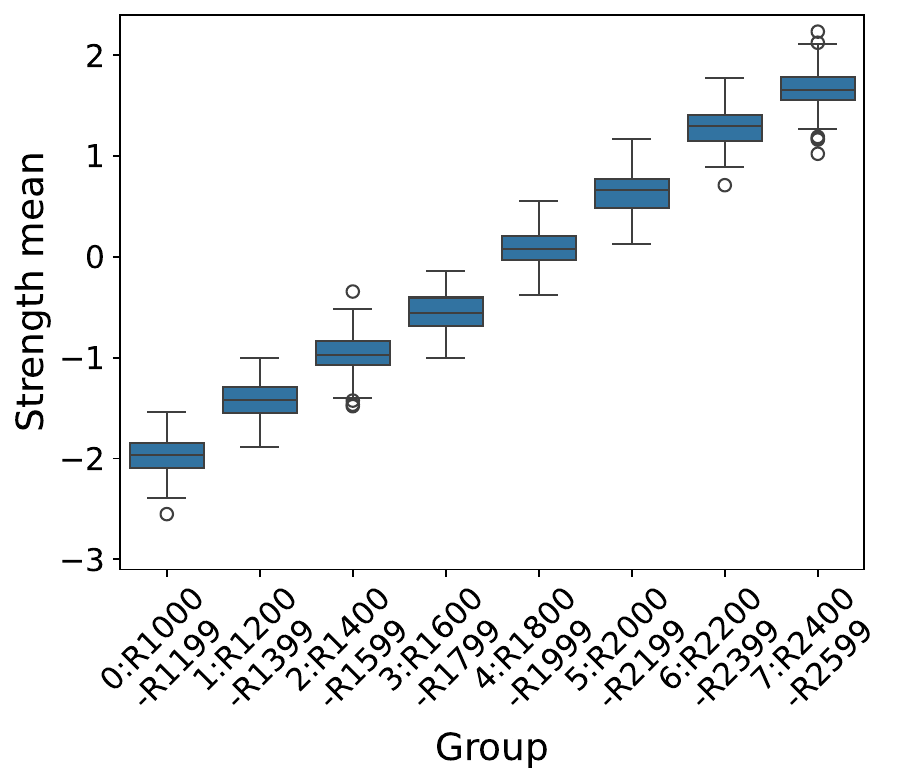}
        \caption{Chess}
        \label{fig:boxplot_str_gmean_chess_random}
    \end{subfigure}
    \caption{The mean strength scores of randomly sampled data points in each rank group with $n=20$.}
    \label{fig:boxplot_str_gmean_random}
\end{figure}

\cref{fig:boxplot_str_gmean_random} shows box plots illustrated in a similar way to \cref{fig:boxplot_str_gmean}.
The difference was that we did random sampling within each rank group instead of player-specific sampling.
In other words, for each rank group, we repeated the following process 100 times: Randomly sample 20 games within the rank groups and obtain the mean strength score.

With random sampling, both Go and chess show significantly less overlap (i.e., narrower distributions), suggesting that rank estimation using only mean strength scores would be more accurate.
As discussed at the beginning of \cref{sec:estimation by player}, the variance in feature distributions becomes smaller because the averaging effects smooth out individual players' characteristics and rating variability.
This effect appears to be particularly strong in chess.
One possible reason is that each rank group's rating range is slightly wider in chess than in Go if converting ranks 5k, \ldots, 9d into ratings.
Additionally, it may be that chess shows greater diversity in representations learned inside the strength estimator.

\section{Conclusion}

In this study, we proposed a novel strength estimation method that improves accuracy by considering the behavior tendency of human players across different skill levels. 
Specifically, in addition to using strength scores obtained from models of a previous state-of-the-art method, we integrated features extracted based on policies from models that imitate human players' gameplay across different skill levels.
We also extracted features related to score losses, which indicate the degree of mistakes made by players when compared to strong AI players' moves.


Experiments on Go and chess showed that our proposed method achieved a higher accuracy compared to a previous state-of-the-art method.
In more detail, our method achieved an accuracy of 87.5\% in Go and 81.5\% in chess when given 15 matches randomly sampled from the same rank group.
The accuracy was 9.3\% and 7.9\% higher than the previous state-of-the-art method in Go and chess, respectively.

Additionally, considering that the strength estimation of individual players better fits practical applications, we conducted experiments under this scenario.
The accuracy was 63.2\% in Go and 60.4\% in chess when 15 matches were sampled from each player, which was 2.1\% and 9.6\% higher than the previous state-of-the-art method, though in all cases, the accuracy was lower than randomly sampling from the same rank group.
These results demonstrated the challenges of player-specific strength estimation compared to group-based estimation, as the averaging effect for the former scenario is weaker than the broader averaging across multiple players in the latter scenario.


Future research directions include exploring the applicability of our method to a wider range of games and different types of data, as well as developing further calibration methods to address outliers and variations in player-specific strength estimation. The proposed approach contributes to teaching or entertaining human players by enabling more accurate player strength estimation, which ultimately enhances human-AI interactions in games.



%
%
%
\bibliographystyle{splncs04}
%

\bibliography{ESML_PKDD}

\newpage
\appendix
\section{Selection Criteria of Game Records}
\label{appendix:selection criteria}

In our experiments, game records of Go were collected from Fox Go~\cite{FoxGo}, and game records of chess were collected from Lichess~\cite{Lichess}.
The selection criteria of the game records were similar to those of Chen et al.~\cite{Chen2025Strength}.

\subsection{Go}

Target players were those who ranked within 3 kyu (3k), \ldots, 1 kyu (1k), 1 dan (1d), 2 dan (2d), \ldots, 8 dan (8d), and 9 dan (9d), from weak to strong.
The players were separated into 11 rank groups: 3--5k, 1--2k, 1d, 2d, \ldots, 8d, and 9d.
For game records of Go, the selection criteria were as follows.
\begin{itemize}
    \item The match must contain at least 50 plies.
    \item The match was ended after both players passed or when one player resigned, instead of disconnection or timeout.
    \item In the match, both players must be in the same rank group.
\end{itemize}

For KataGo HumanSL~\cite{UrlKataGoV15}, two sets of rank settings can be used to model amateur players.
Specifically, to obtain a specific policy for an amateur rank $x$, one can set the variable \texttt{humanSLProfile} in the configuration file to either \texttt{preaz\_$x$} or \texttt{rank\_$x$}.
\texttt{preaz\_$x$} was used to model human players' moves before AlphaZero's opening became popular, while \texttt{rank\_$x$} was used to model those after the openings were influenced by AlphaZero.
Since game records collected from the Fox Go dataset~\cite{FoxGo} were played around 2017, the time before AlphaZero's opening became popular.
Thus, when using KataGo HumanSL to obtain policies across different ranks, we used the \texttt{preaz\_$x$} settings.

\subsection{Chess}

Target players were those whose ratings were from R1000 to R2599, from weak to strong.
The players were separated into 8 rank groups: R1000--R1199, R1200--R1399, \ldots, R2200--R2399, and R2400--R2599.
For game records of chess, the selection criteria were as follows.
\begin{itemize}
    \item The match was played in February 2024.
    \item The match must contain at least 20 plies.
    \item The time control of the match was blitz.
    \item In the match, both players must be in the same rank group.
\end{itemize}


\section{Confusion Matrices of Rank Estimation}
\label{appendix:confusion matrix group}

\begin{figure}[bt]
   \centering
    \begin{subfigure}[t]{0.49\linewidth}
        \centering
        \raisebox{2mm}{\includegraphics[width=43mm]{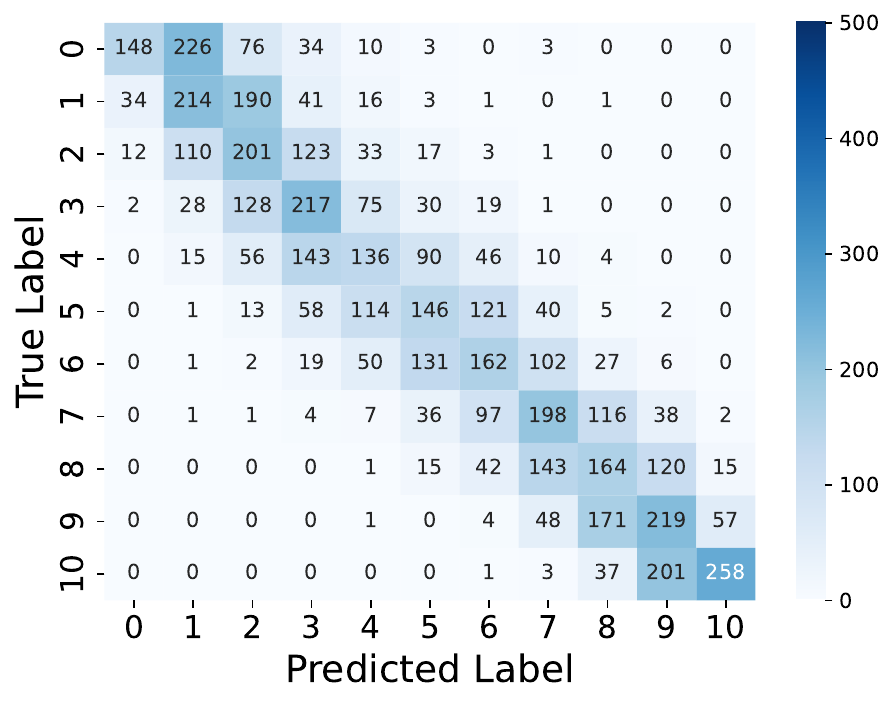}}
        \caption{Proposed method in Go}
        \vspace*{1.5mm}
        \label{}
    \end{subfigure}
    \begin{subfigure}[t]{0.49\linewidth}
        \centering
        \includegraphics[width=43mm]{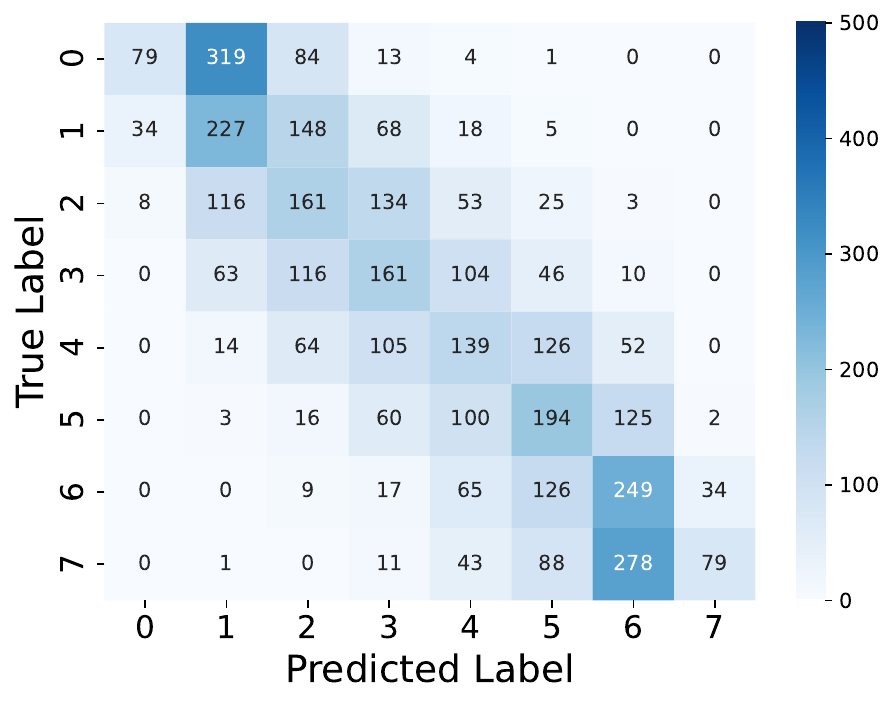}
        \caption{Proposed method in chess}
        \vspace*{1.5mm}
        \label{}
    \end{subfigure}
    \begin{subfigure}[t]{0.49\linewidth}
        \centering
        \raisebox{2mm}{\includegraphics[width=43mm]{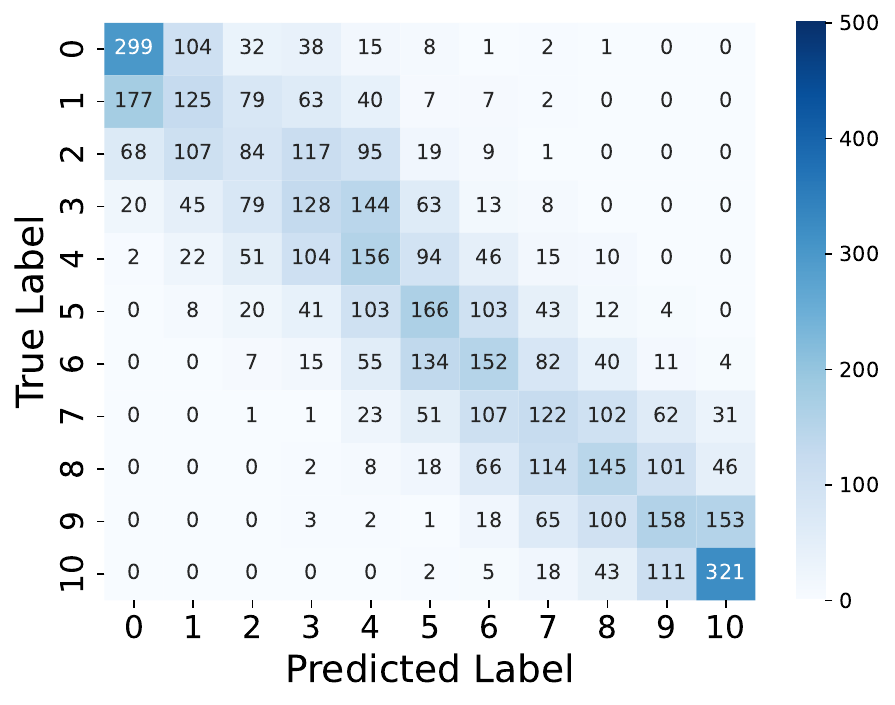}}
        \caption{Chen et al.'s method~\cite{Chen2025Strength} in Go}
        \label{}
    \end{subfigure}
    \begin{subfigure}[t]{0.49\linewidth}
        \centering
        \includegraphics[width=43mm]{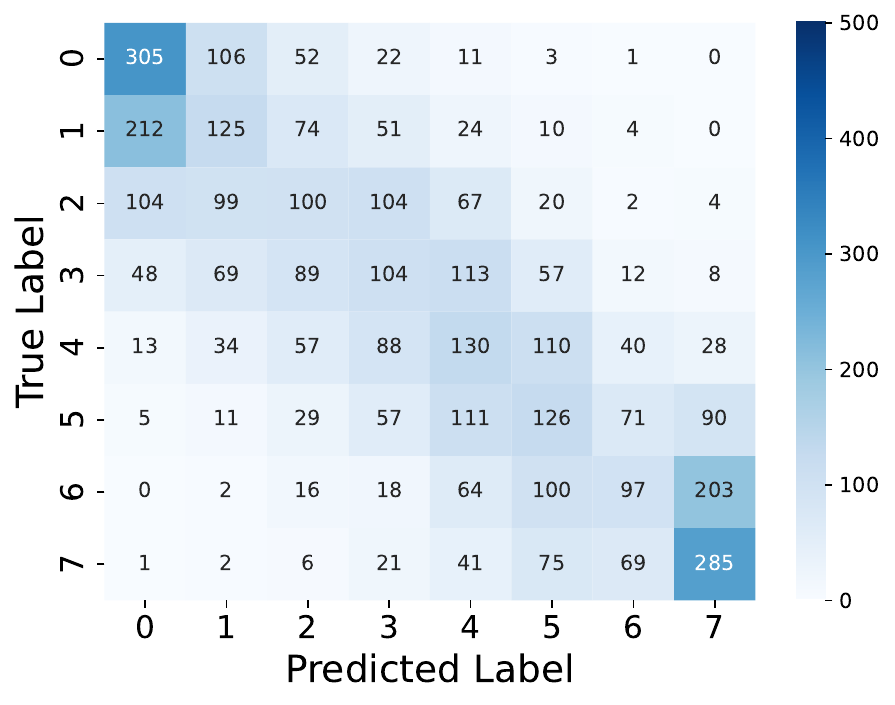}
        \caption{Chen et al.'s method~\cite{Chen2025Strength} in chess}
        \label{}
    \end{subfigure}
    \caption{Confusion matrices of rank estimation with $n=1$.}
    \label{fig:confusion matrix group n1}
\end{figure}

\begin{figure}[bt]
   \centering
    \begin{subfigure}[t]{0.49\linewidth}
        \centering
        \raisebox{2mm}{\includegraphics[width=43mm]{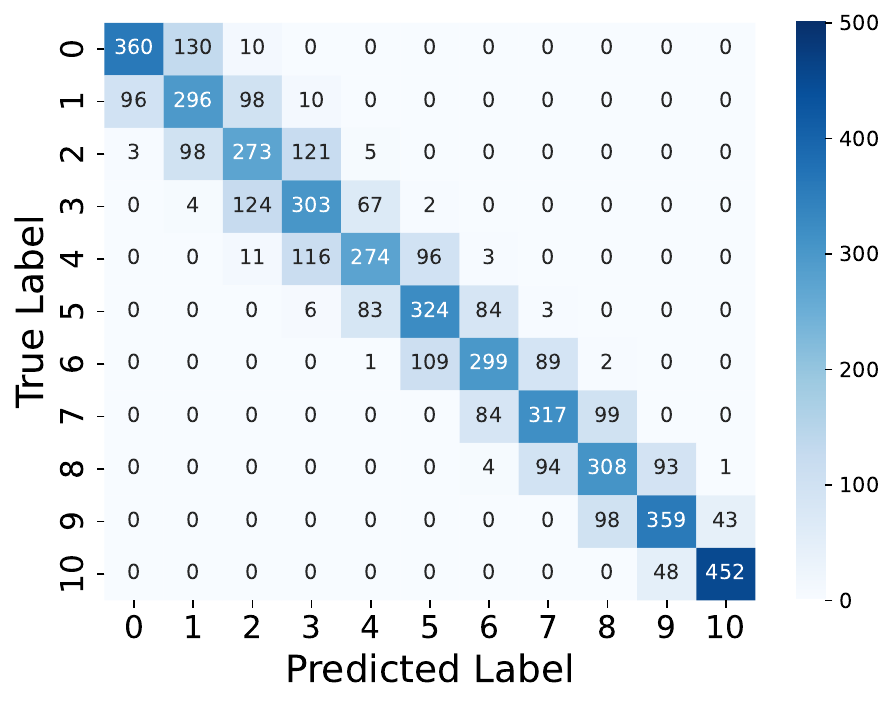}}
        \caption{Proposed method in Go}
        \vspace*{1.5mm}
        \label{}
    \end{subfigure}
    \begin{subfigure}[t]{0.49\linewidth}
        \centering
        \includegraphics[width=43mm]{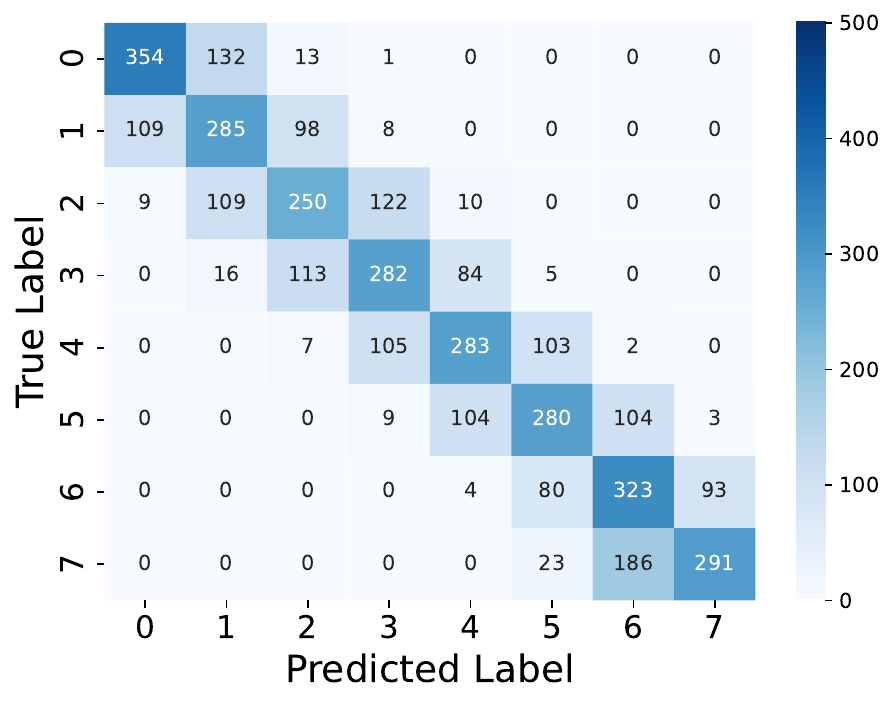}
        \caption{Proposed method in chess}
        \vspace*{1.5mm}
        \label{}
    \end{subfigure}
    \begin{subfigure}[t]{0.49\linewidth}
        \centering
        \raisebox{2mm}{\includegraphics[width=43mm]{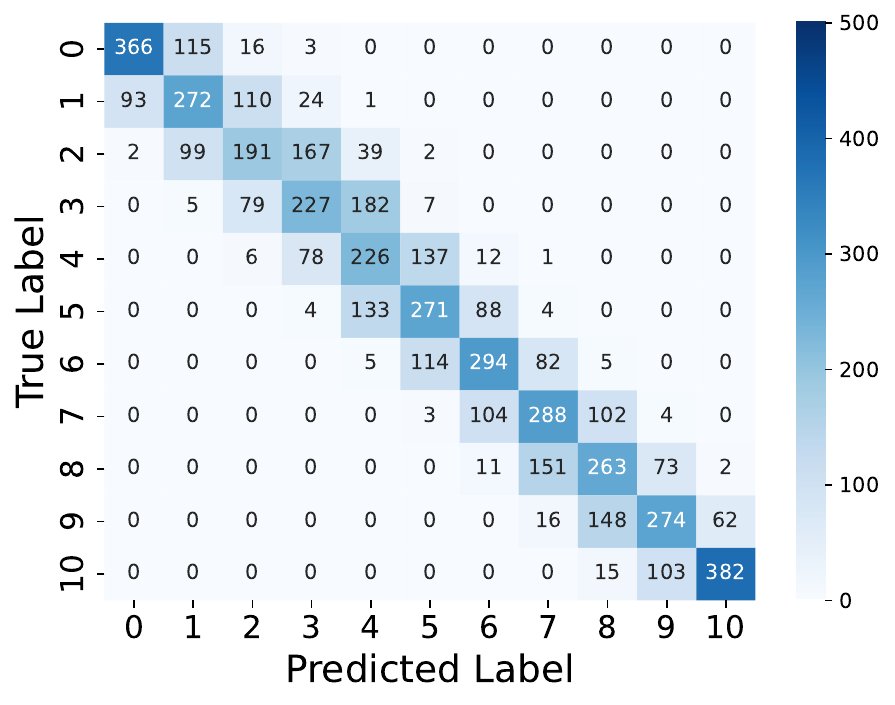}}
        \caption{Chen et al.'s method~\cite{Chen2025Strength} in Go}
        \label{}
    \end{subfigure}
    \begin{subfigure}[t]{0.49\linewidth}
        \centering
        \includegraphics[width=43mm]{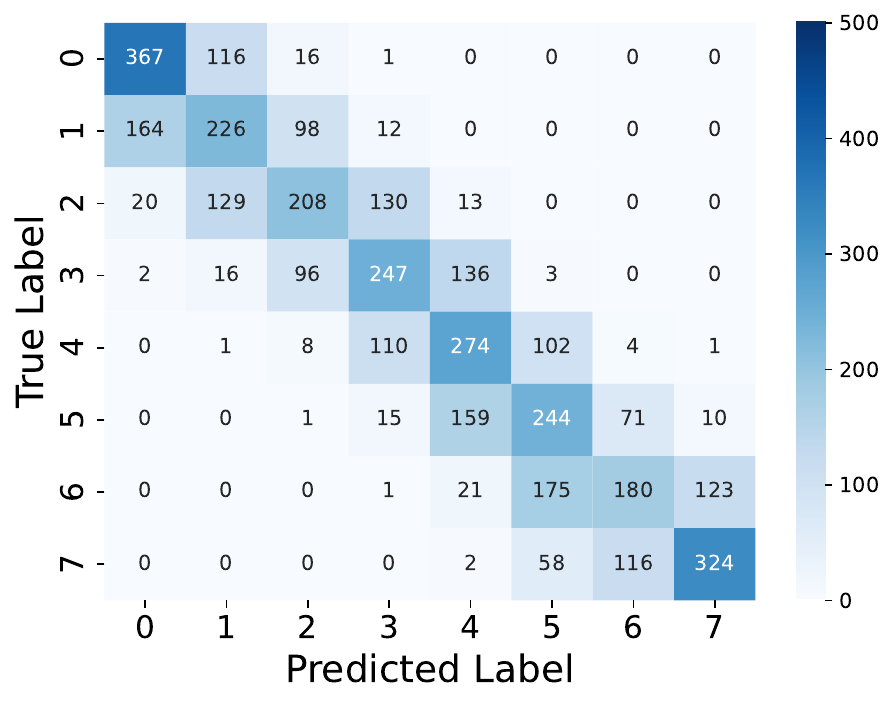}
        \caption{Chen et al.'s method~\cite{Chen2025Strength} in chess}
        \label{}
    \end{subfigure}
    \caption{Confusion matrices of rank estimation with $n=5$.}
    \label{fig:confusion matrix group n5}
\end{figure}

\begin{figure}[bt]
   \centering
    \begin{subfigure}[t]{0.49\linewidth}
        \centering
        \raisebox{2mm}{\includegraphics[width=43mm]{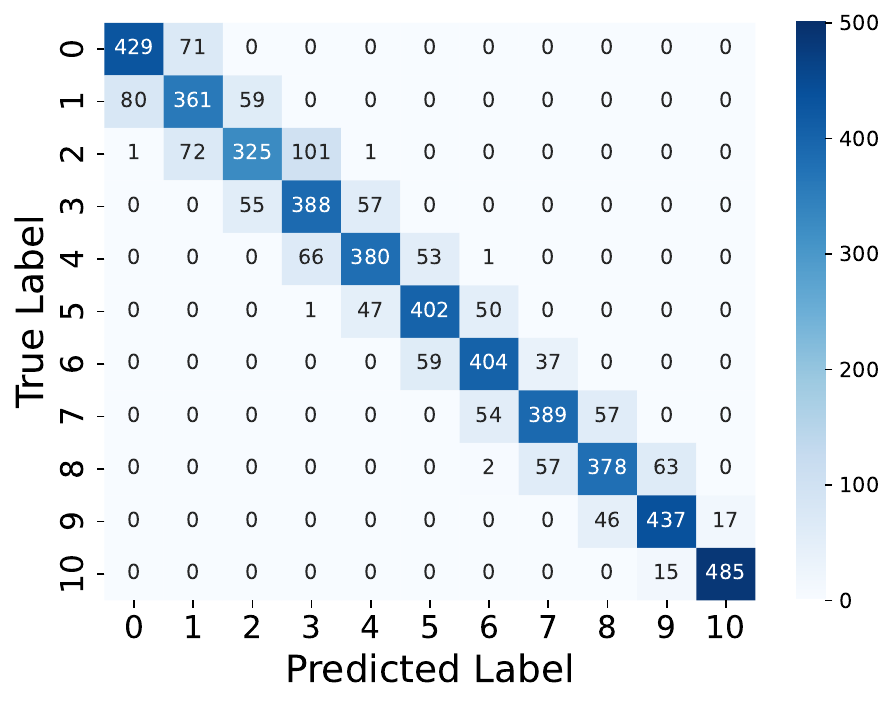}}
        \caption{Proposed method in Go}
        \vspace*{1.5mm}
        \label{}
    \end{subfigure}
    \begin{subfigure}[t]{0.49\linewidth}
        \centering
        \includegraphics[width=43mm]{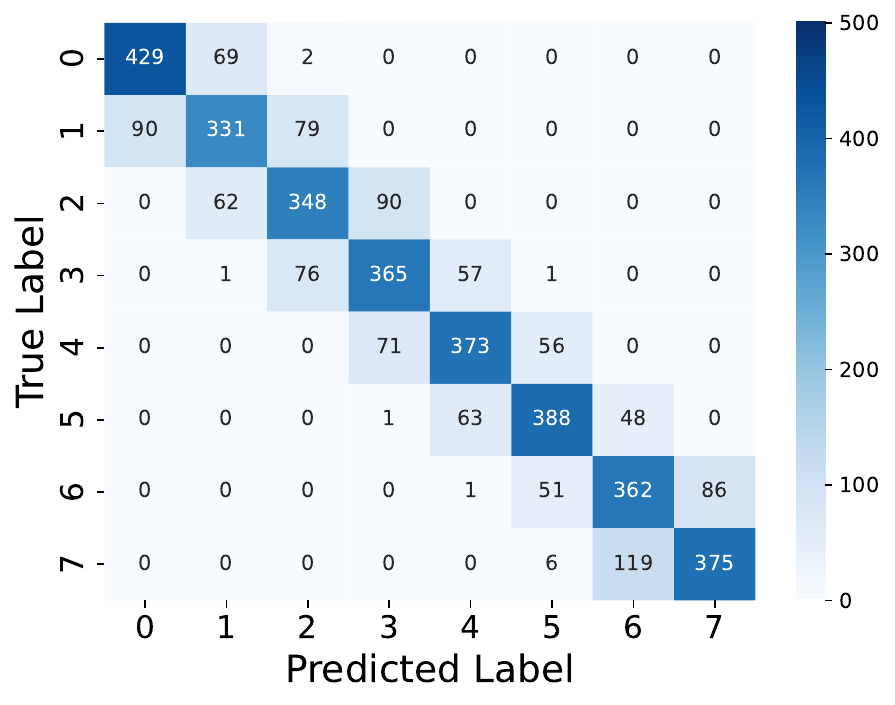}
        \caption{Proposed method in chess}
        \vspace*{1.5mm}
        \label{}
    \end{subfigure}
    \begin{subfigure}[t]{0.49\linewidth}
        \centering
        \raisebox{2mm}{\includegraphics[width=43mm]{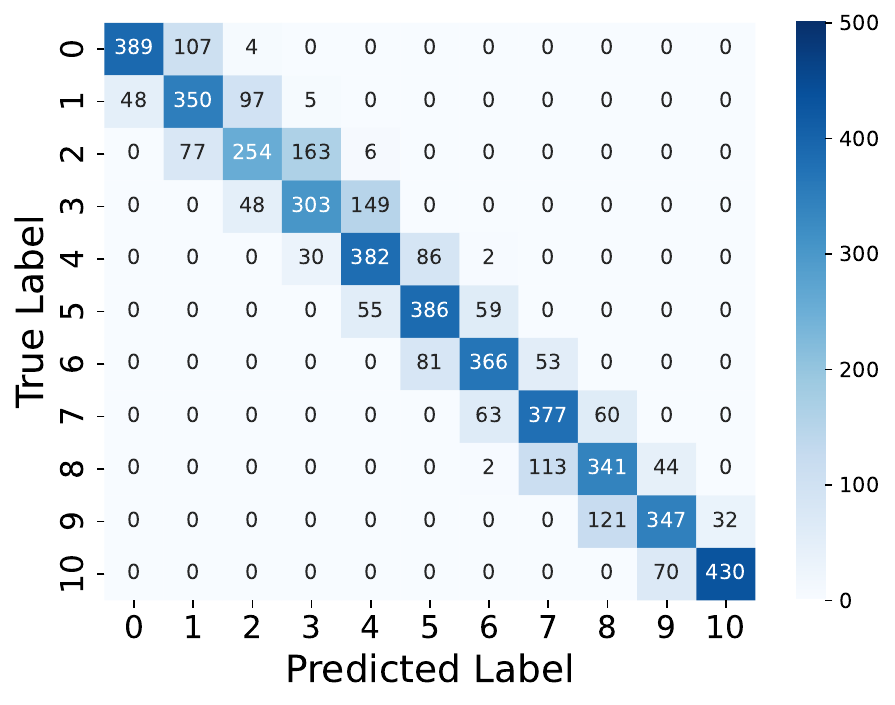}}
        \caption{Chen et al.'s method~\cite{Chen2025Strength} in Go}
        \label{}
    \end{subfigure}
    \begin{subfigure}[t]{0.49\linewidth}
        \centering
        \includegraphics[width=43mm]{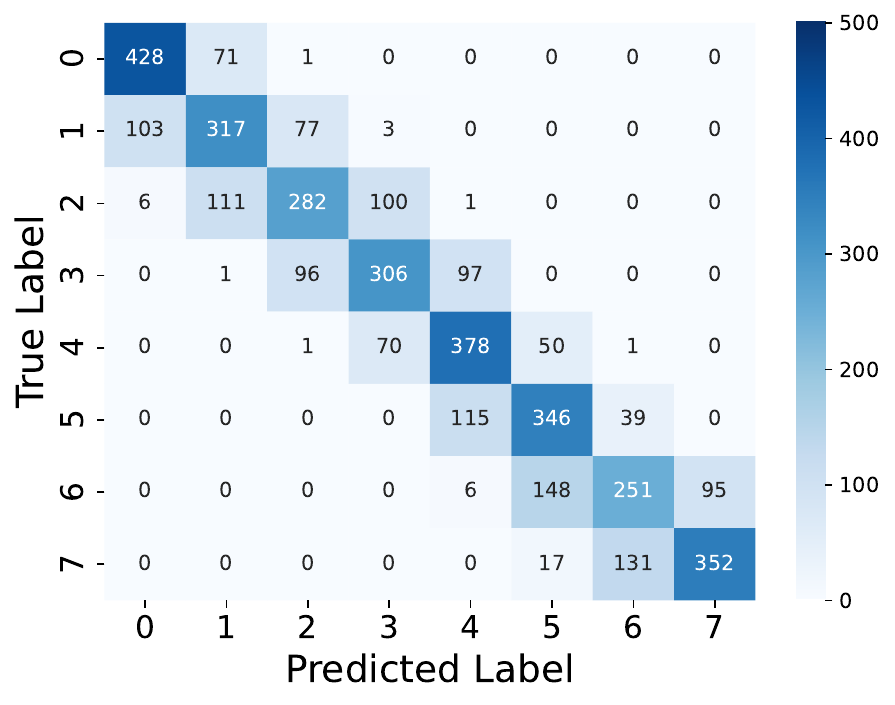}
        \caption{Chen et al.'s method~\cite{Chen2025Strength} in chess}
        \label{}
    \end{subfigure}
    \caption{Confusion matrices of rank estimation with $n=10$.}
    \label{fig:confusion matrix group n10}
\end{figure}

\begin{figure}[bt]
   \centering
    \begin{subfigure}[t]{0.49\linewidth}
        \centering
        \raisebox{2mm}{\includegraphics[width=43mm]{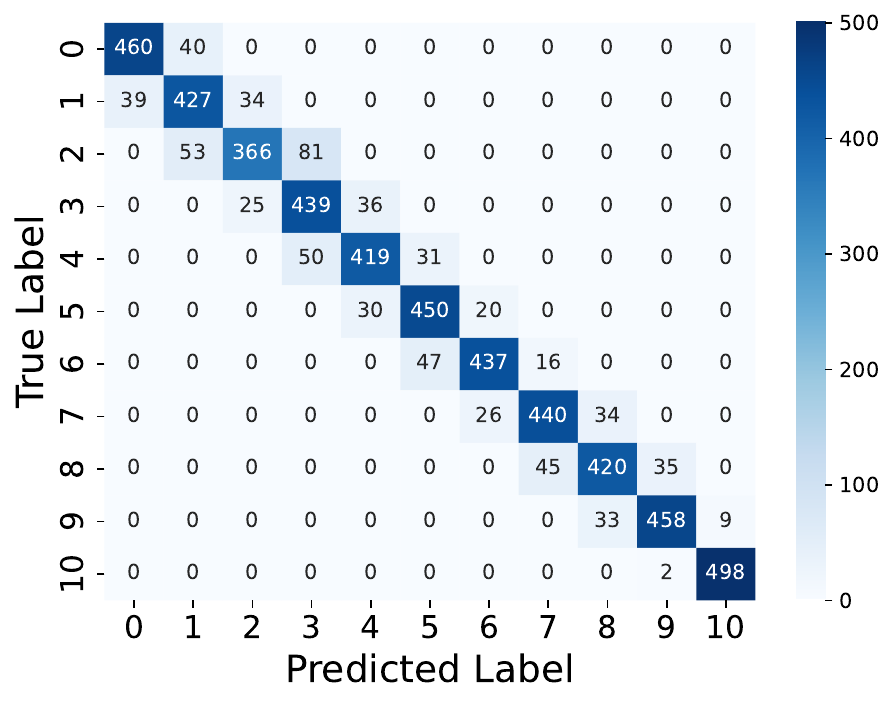}}
        \caption{Proposed method in Go}
        \vspace*{1.5mm}
        \label{}
    \end{subfigure}
    \begin{subfigure}[t]{0.49\linewidth}
        \centering
        \includegraphics[width=43mm]{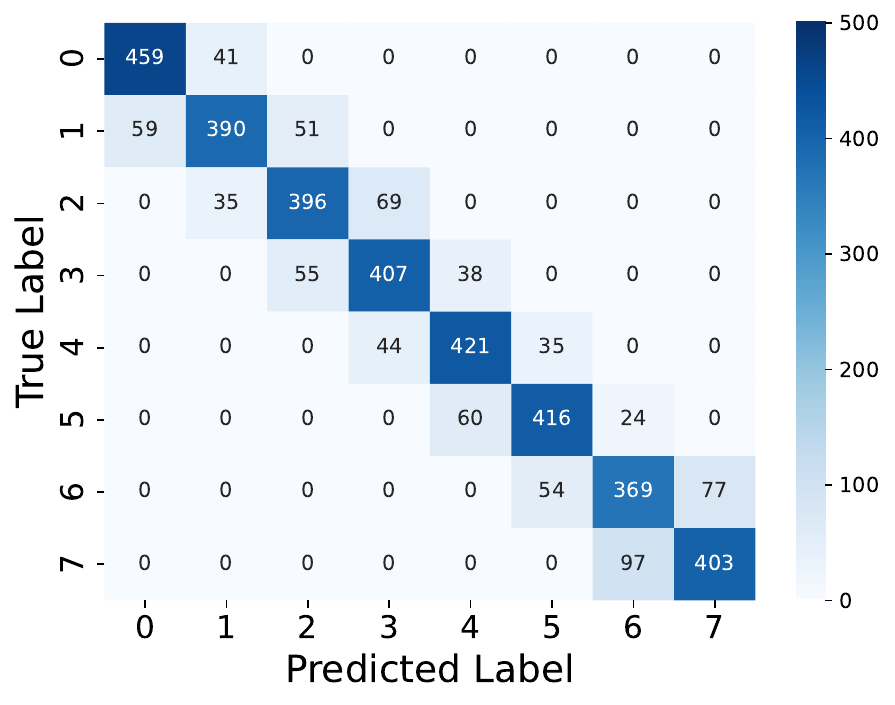}
        \caption{Proposed method in chess}
        \vspace*{1.5mm}
        \label{}
    \end{subfigure}
    \begin{subfigure}[t]{0.49\linewidth}
        \centering
        \raisebox{2mm}{\includegraphics[width=43mm]{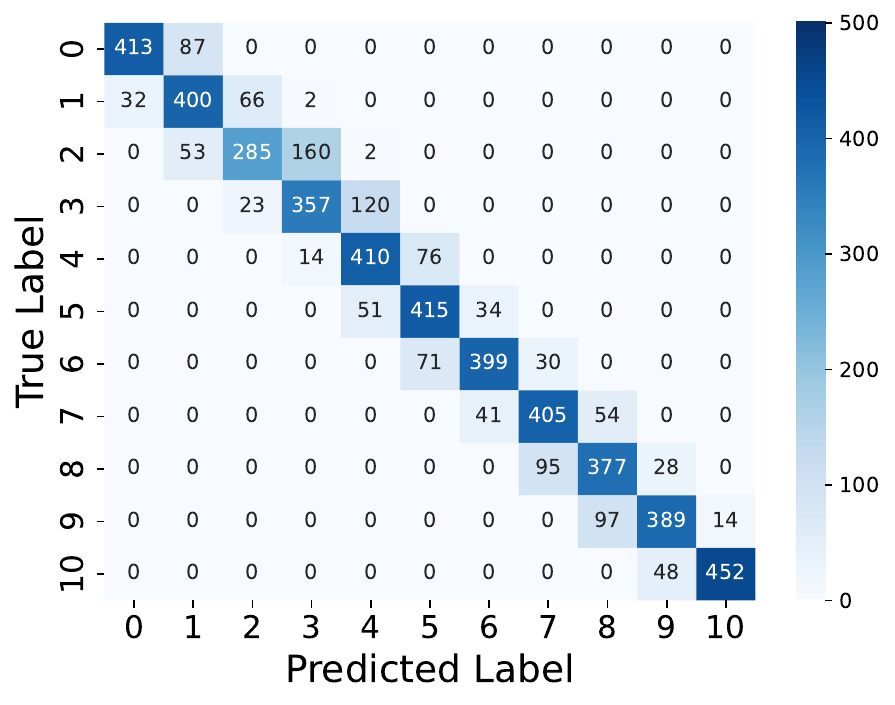}}
        \caption{Chen et al.'s method~\cite{Chen2025Strength} in Go}
        \label{}
    \end{subfigure}
    \begin{subfigure}[t]{0.49\linewidth}
        \centering
        \includegraphics[width=43mm]{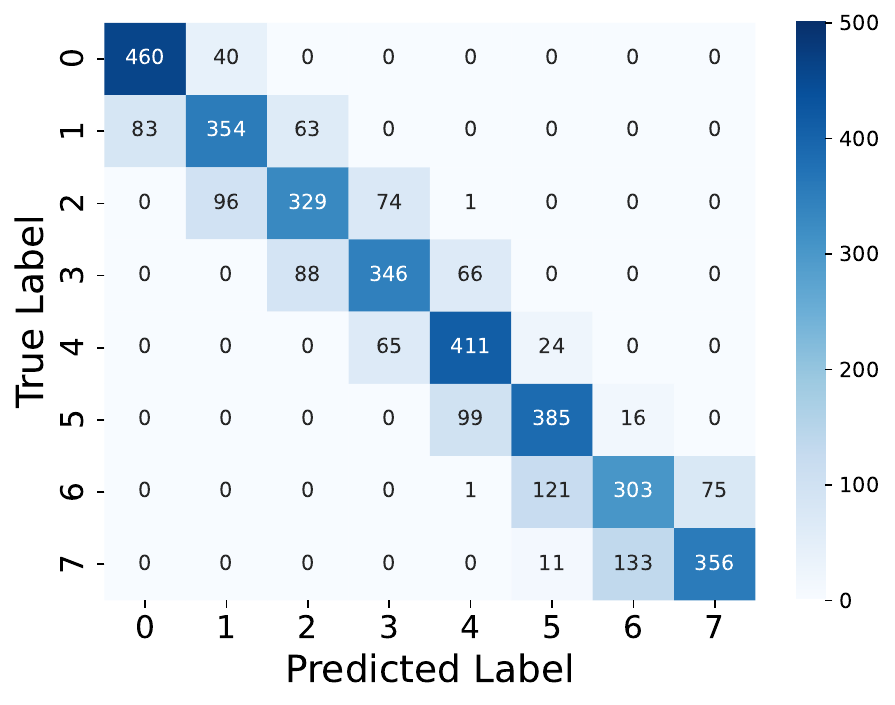}
        \caption{Chen et al.'s method~\cite{Chen2025Strength} in chess}
        \label{}
    \end{subfigure}
    \caption{Confusion matrices of rank estimation with $n=15$.}
    \label{fig:confusion matrix group n15}
\end{figure}

\begin{figure}[bt]
   \centering
    \begin{subfigure}[t]{0.49\linewidth}
        \centering
        \raisebox{2mm}{\includegraphics[width=43mm]{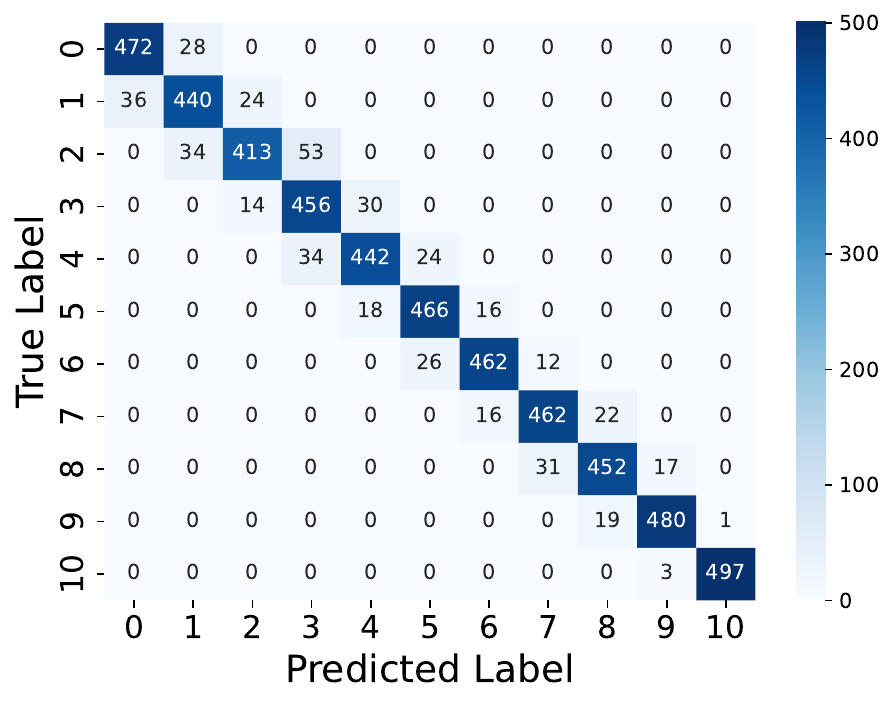}}
        \caption{Proposed method in Go}
        \vspace*{1.5mm}
        \label{}
    \end{subfigure}
    \begin{subfigure}[t]{0.49\linewidth}
        \centering
        \includegraphics[width=43mm]{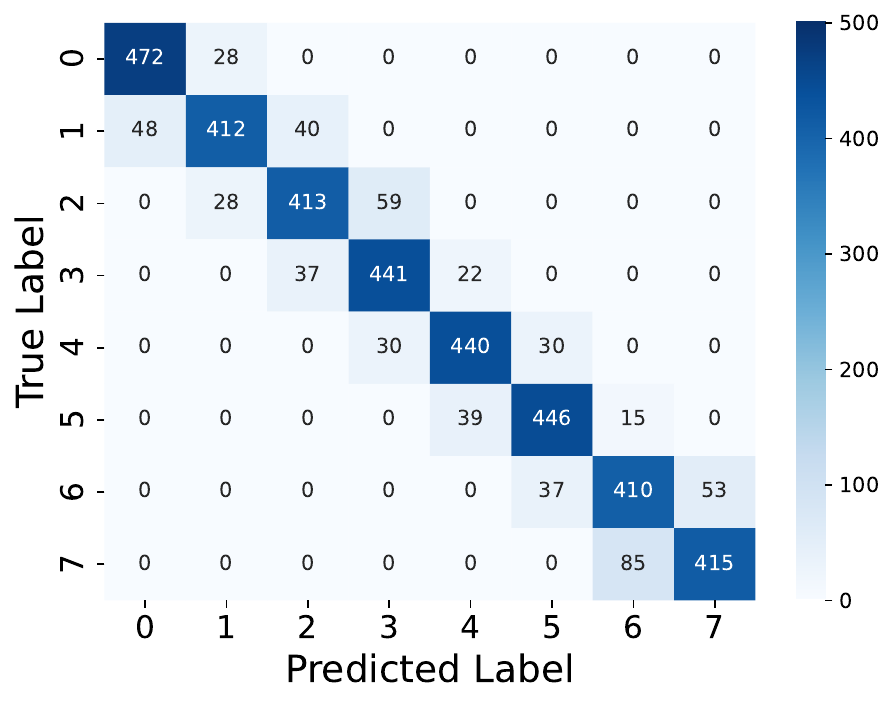}
        \caption{Proposed method in chess}
        \vspace*{1.5mm}
        \label{}
    \end{subfigure}
    \begin{subfigure}[t]{0.49\linewidth}
        \centering
        \raisebox{2mm}{\includegraphics[width=43mm]{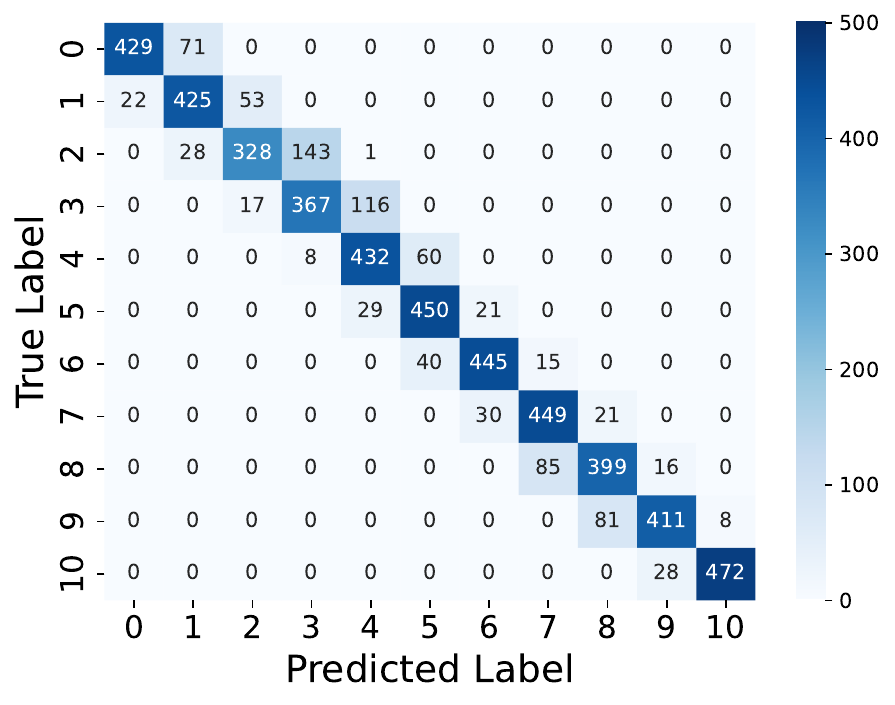}}
        \caption{Chen et al.'s method~\cite{Chen2025Strength} in Go}
        \label{}
    \end{subfigure}
    \begin{subfigure}[t]{0.49\linewidth}
        \centering
        \includegraphics[width=43mm]{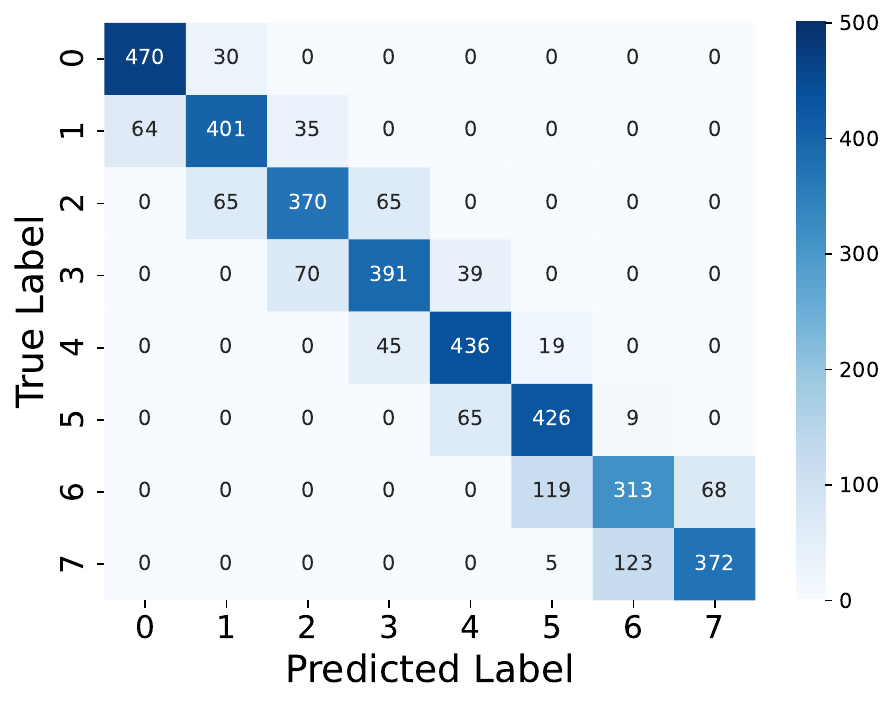}
        \caption{Chen et al.'s method~\cite{Chen2025Strength} in chess}
        \label{}
    \end{subfigure}
    \caption{Confusion matrices of rank estimation with $n=20$.}
    \label{fig:confusion matrix group n20}
\end{figure}

\cref{fig:confusion matrix group n1,fig:confusion matrix group n5,fig:confusion matrix group n10,fig:confusion matrix group n15,fig:confusion matrix group n20} show the confusion matrix of rank estimation for both our proposed method and Chen et al.'s method~\cite{Chen2025Strength} in both Go and chess with $n\in \{1, 5, 10, 15, 20\}$.
Namely, these confusion matrices present more detailed results of \cref{tab:result1}.
The results showed that both methods performed well, with most predictions concentrated around $x=y$.
While Chen et al.'s method predicted better at the two ends, i.e., groups 0 and 10 in Go and groups 0 and 7 in chess, our method predicted better in the middle groups (darker colors on $x=y$).
This indicates that our method achieved more consistent performance across different rank groups, which leads to improved accuracy.

\clearpage

\section{Role Division in Imitation Models}
\label{appendix:maia1500 maia1900}

In \cref{subsec:discussions main}, we assumed that different imitation models are useful to distinguish different rank groups.
For example, the Maia-1900 model might be useful to distinguish players within rank groups 0 to 3 (R1000 to R1799), but not rank groups 4 to 7 (R1800 to R2599).
The Maia-1500 model might be the reverse.
To support the above hypothesis, we conducted a simplified experiment, where we only used features extracted from Maia models.
Specifically, we compared three settings: ``Maia-1500 only'', ``Maia-1900 only'', and ``both''.

\begin{table}[bt]
    \centering
    \caption{Accuracy of rank estimation in chess with $n=20$ using only Maia-1500, Maia-1900, or both}
    \label{tab:maia_only_chess}
    \begin{tabular}{c|cccccccc|c}
        \toprule
         & $g_0$ & $g_1$ & $g_2$ & $g_3$ & $g_4$ & $g_5$ & $g_6$ & $g_7$ & Overall \\
        \midrule
        Maia-1500 Only & 0.0 & 0.0 & 0.2 & \textbf{87.4} & 23.0 & 0.0 & 0.0 & 0.0   & 13.8\\
        Maia-1900 Only & 61.4 & 54.8 & 27.2 & 15.6 & 18.8 & \textbf{85.8} & 0.0 & 0.0 & 33.0\\
        \midrule
        Both & \textbf{78.4} & \textbf{64.8} & \textbf{59.6} & 60.2  & \textbf{70.0} & 57.6 & \textbf{72.2} & \textbf{57.8} & \textbf{65.1}\\
        \bottomrule
    \end{tabular}
\end{table}

\cref{tab:maia_only_chess} shows the estimation accuracy of each rank group in chess.
When using only Maia-1900, the rank estimation model obtained relatively high accuracy for low-rated groups such as 0 and 1, where Maia-1900 had low prior geometric means.
For rank groups 4 to 7, where the prior geometric means were close, the rank estimation model predicted almost all players to be rank group 5.
When using only Maia-1500, the accuracy was relatively high for rank groups 3 and 4, where the prior geometric means were relatively high.
For rank groups 0, 1, 6, and 7, the prior geometric means were relatively low, and the rank estimation model could not distinguish whether the unlikeness was because the player was stronger or weaker.
In contrast, when using both, the rank estimation model could utilize the pattern that Maia-1500 showed a monotonic decrease in the prior geometric means in rank groups 4 to 7, where Maia-1900 had high prior geometric means.
This allowed the rank estimation model to infer that a lower prior geometric mean with Maia-1500 in this range indicated a stronger player, leading to a significant improvement in overall accuracy.

\section{Confusion Matrices of Player-Specific Rank Estimation}
\label{appendix:confusion matrix player}

\begin{figure}[bt]
   \centering
    \begin{subfigure}[t]{0.49\linewidth}
        \centering
        \raisebox{2mm}{\includegraphics[width=43mm]{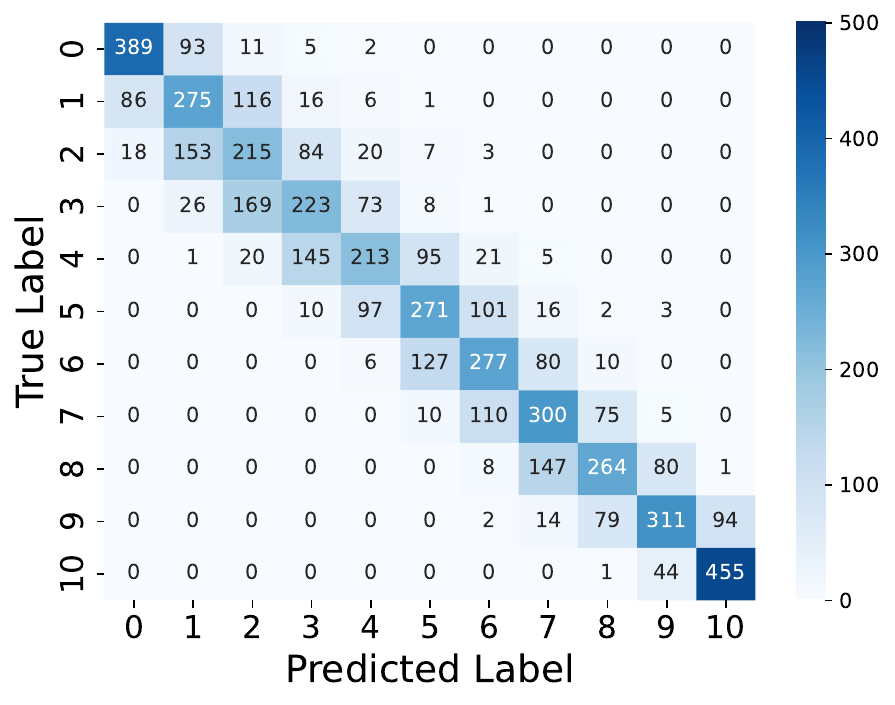}}
        \caption{Proposed method in Go}
        \vspace*{1.5mm}
        \label{}
    \end{subfigure}
    \begin{subfigure}[t]{0.49\linewidth}
        \centering
        \includegraphics[width=43mm]{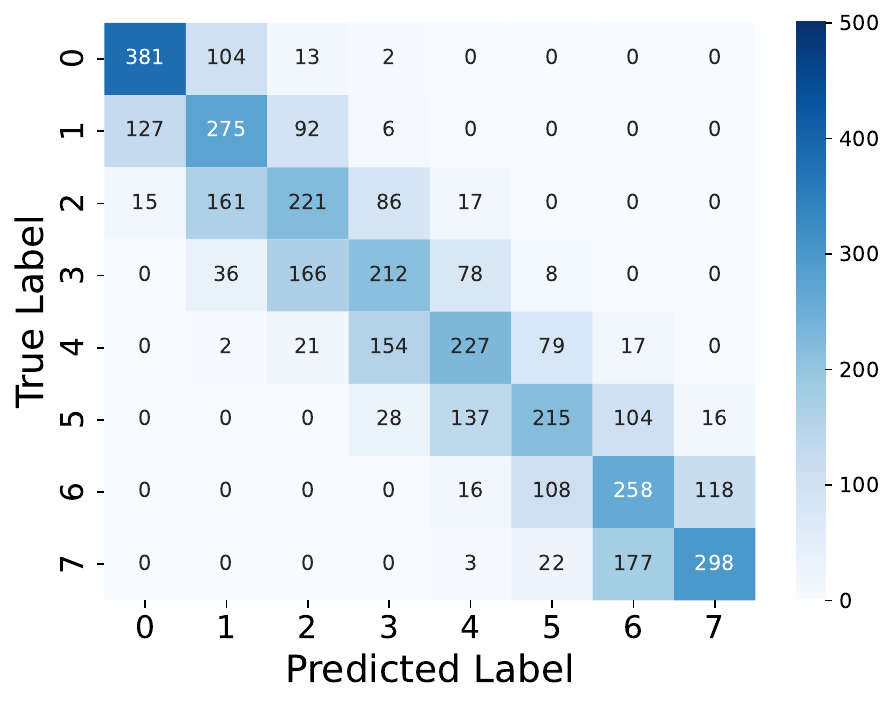}
        \caption{Proposed method in chess}
        \vspace*{1.5mm}
        \label{}
    \end{subfigure}
    \begin{subfigure}[t]{0.49\linewidth}
        \centering
        \raisebox{2mm}{\includegraphics[width=43mm]{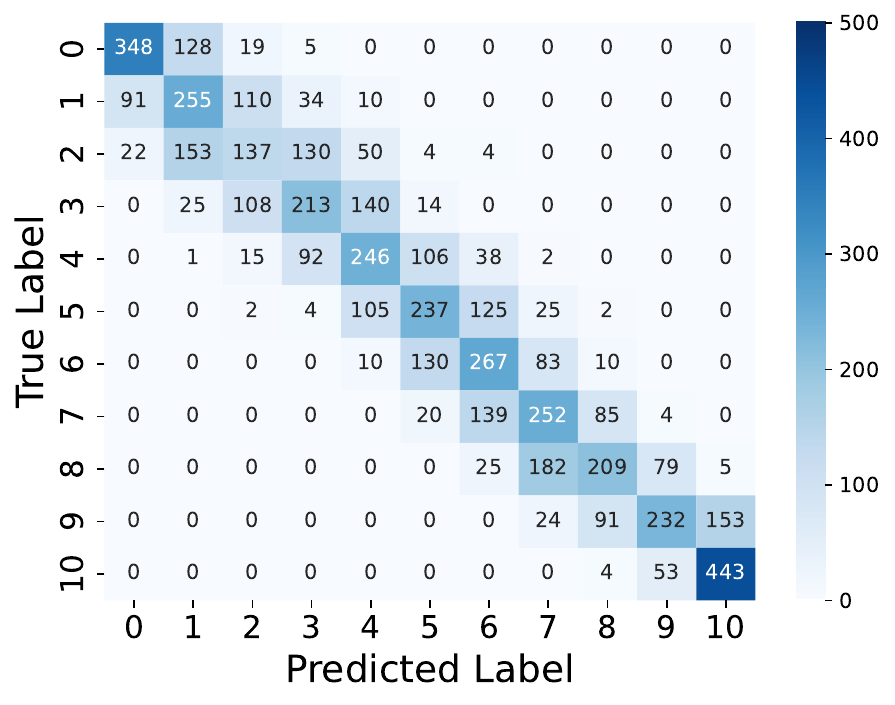}}
        \caption{Chen et al.'s method~\cite{Chen2025Strength} in Go}
        \label{}
    \end{subfigure}
    \begin{subfigure}[t]{0.49\linewidth}
        \centering
        \includegraphics[width=43mm]{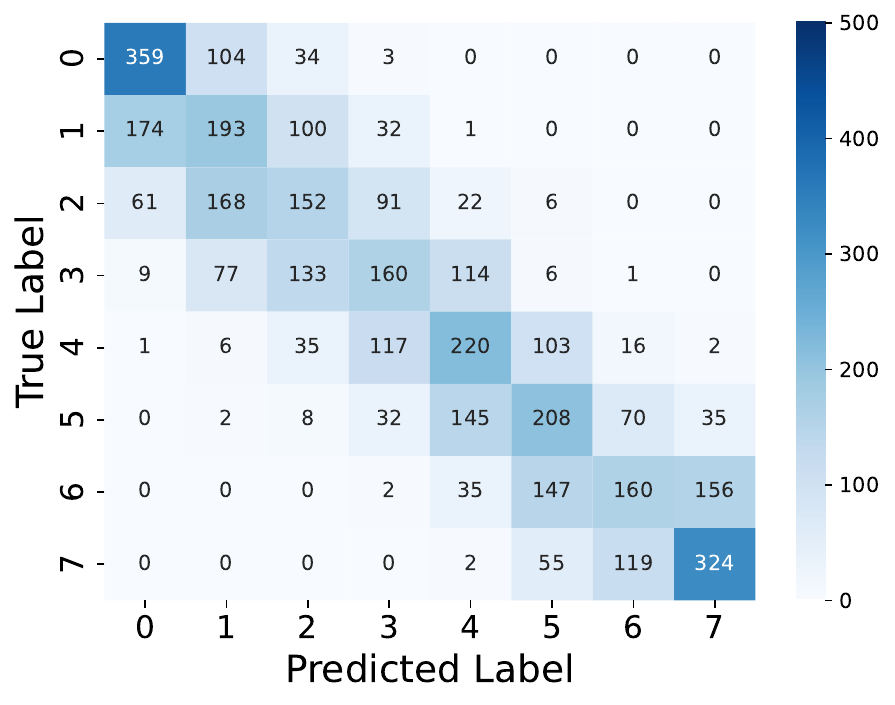}
        \caption{Chen et al.'s method~\cite{Chen2025Strength} in chess}
        \label{}
    \end{subfigure}
    \caption{Confusion matrices of player-specific rank estimation with $n=5$.}
    \label{fig:confusion matrix player n5}
\end{figure}

\begin{figure}[bt]
   \centering
    \begin{subfigure}[t]{0.49\linewidth}
        \centering
        \raisebox{2mm}{\includegraphics[width=43mm]{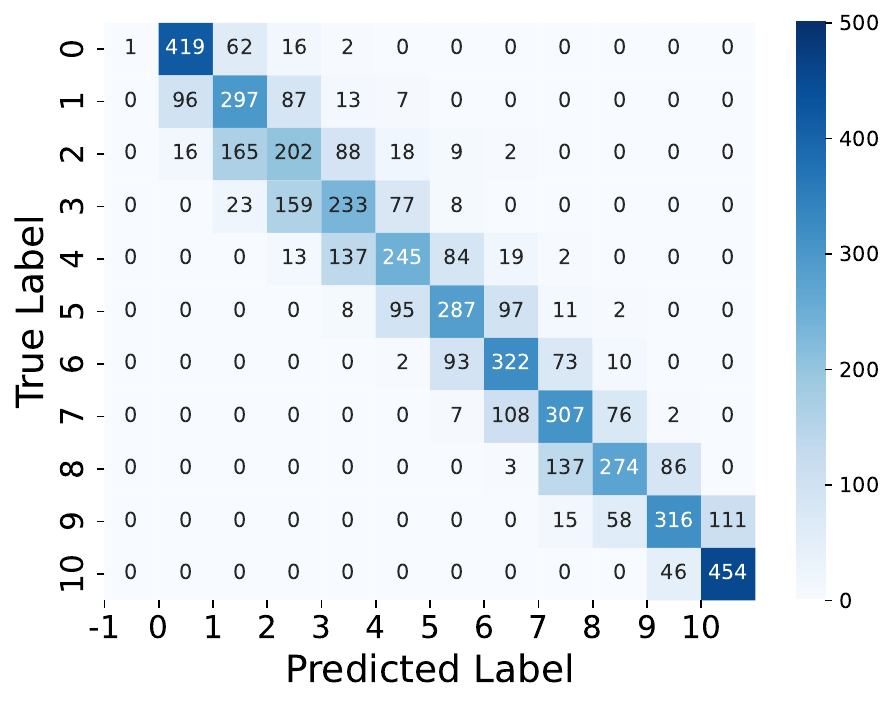}}
        \caption{Proposed method in Go}
        \vspace*{1.5mm}
        \label{}
    \end{subfigure}
    \begin{subfigure}[t]{0.49\linewidth}
        \centering
        \includegraphics[width=43mm]{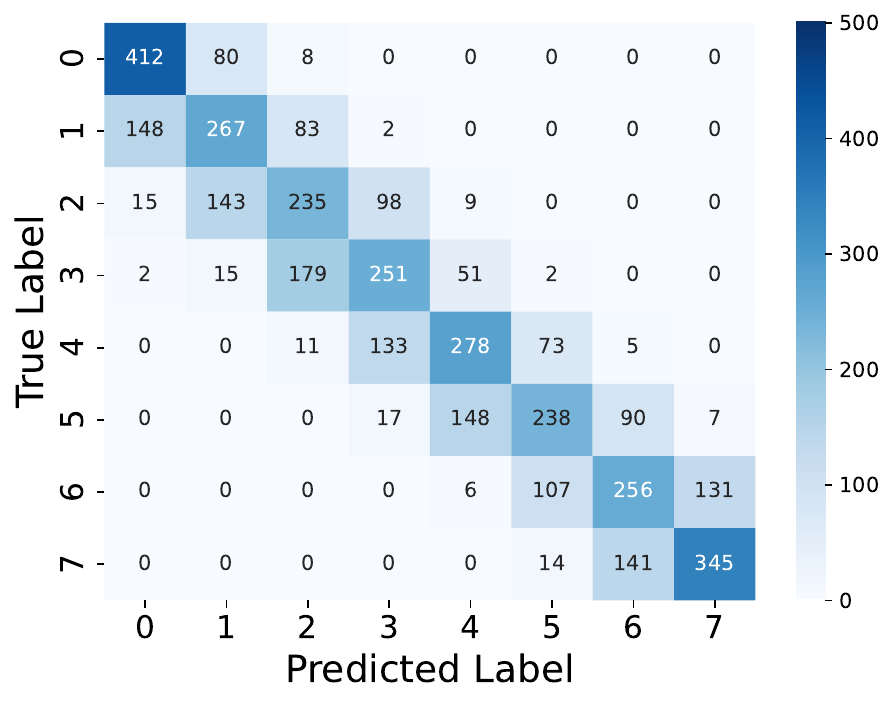}
        \caption{Proposed method in chess}
        \vspace*{1.5mm}
        \label{}
    \end{subfigure}
    \begin{subfigure}[t]{0.49\linewidth}
        \centering
        \raisebox{2mm}{\includegraphics[width=43mm]{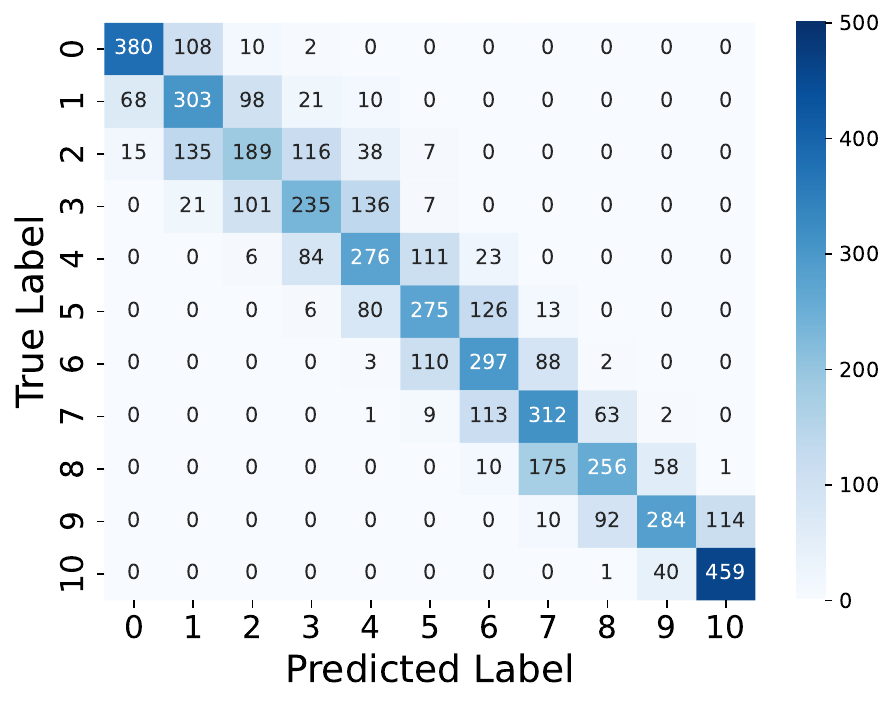}}
        \caption{Chen et al.'s method~\cite{Chen2025Strength} in Go}
        \label{}
    \end{subfigure}
    \begin{subfigure}[t]{0.49\linewidth}
        \centering
        \includegraphics[width=43mm]{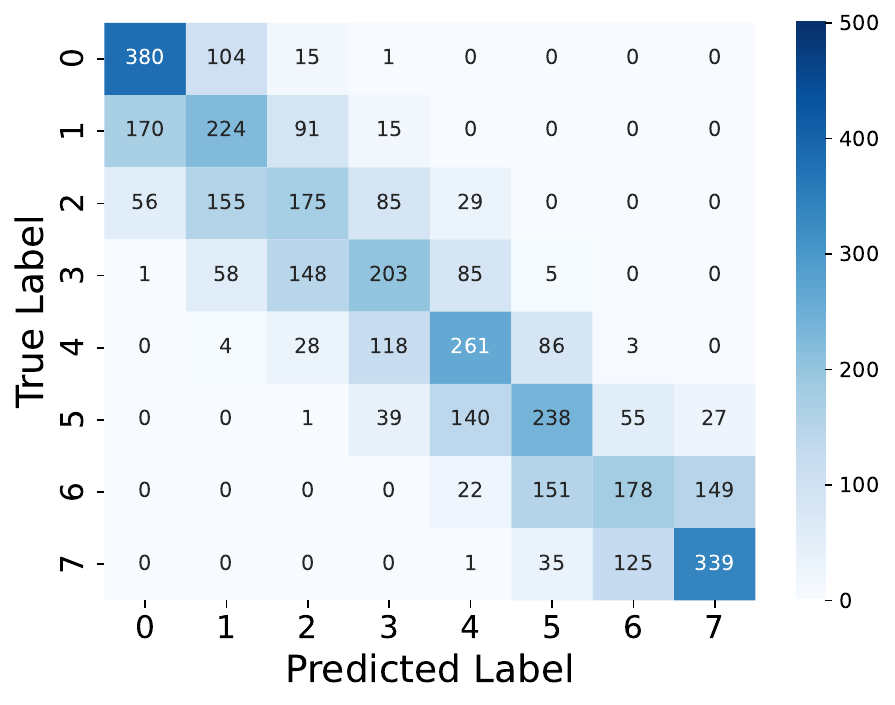}
        \caption{Chen et al.'s method~\cite{Chen2025Strength} in chess}
        \label{}
    \end{subfigure}
    \caption{Confusion matrices of player-specific rank estimation with $n=10$.}
    \label{fig:confusion matrix player n10}
\end{figure}

\begin{figure}[bt]
   \centering
    \begin{subfigure}[t]{0.49\linewidth}
        \centering
        \raisebox{2mm}{\includegraphics[width=43mm]{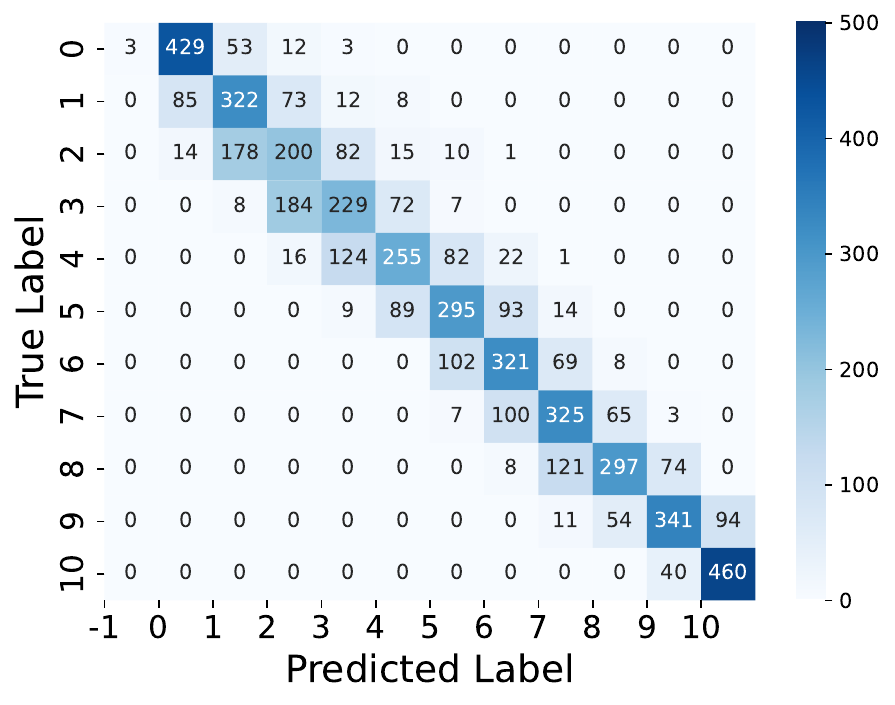}}
        \caption{Proposed method in Go}
        \vspace*{1.5mm}
        \label{}
    \end{subfigure}
    \begin{subfigure}[t]{0.49\linewidth}
        \centering
        \includegraphics[width=43mm]{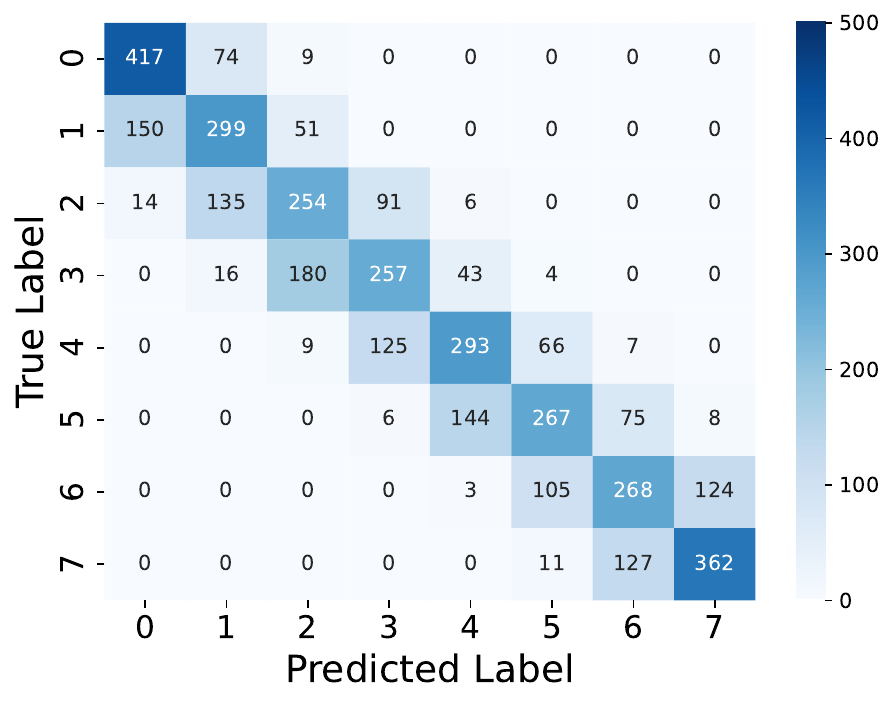}
        \caption{Proposed method in chess}
        \vspace*{1.5mm}
        \label{}
    \end{subfigure}
    \begin{subfigure}[t]{0.49\linewidth}
        \centering
        \raisebox{2mm}{\includegraphics[width=43mm]{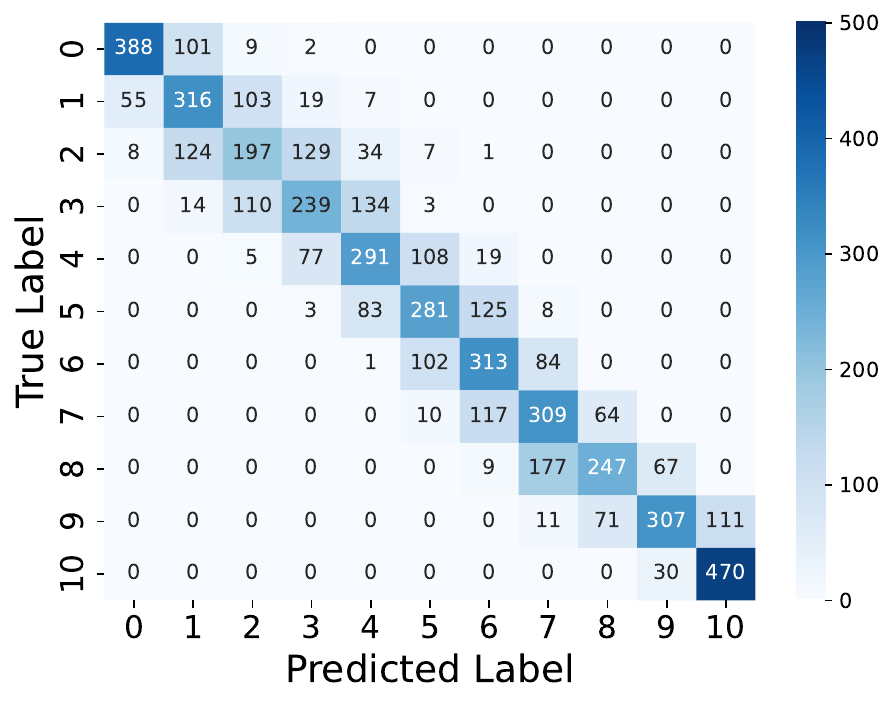}}
        \caption{Chen et al.'s method~\cite{Chen2025Strength} in Go}
        \label{}
    \end{subfigure}
    \begin{subfigure}[t]{0.49\linewidth}
        \centering
        \includegraphics[width=43mm]{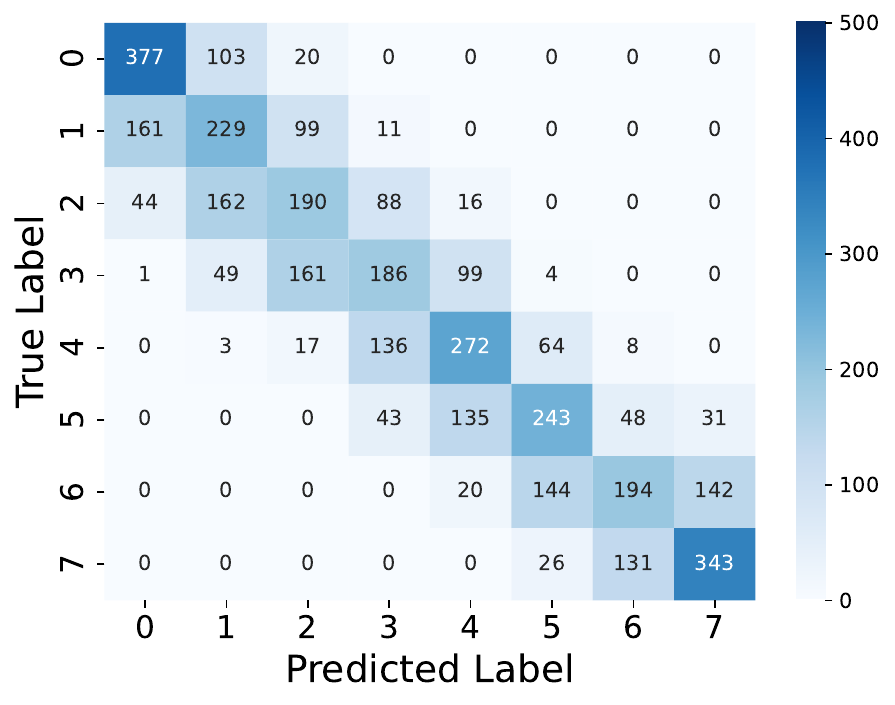}
        \caption{Chen et al.'s method~\cite{Chen2025Strength} in chess}
        \label{}
    \end{subfigure}
    \caption{Confusion matrices of player-specific rank estimation with $n=15$.}
    \label{fig:confusion matrix player n15}
\end{figure}

\cref{fig:confusion matrix player n5,fig:confusion matrix player n10,fig:confusion matrix player n15} show the confusion matrix of player-specific rank estimation for both our proposed method and Chen et al.'s method~\cite{Chen2025Strength} in both Go and chess with $n\in \{5, 10, 15\}$.
Namely, these confusion matrices present more detailed results of the ``Player-Specific Sampling'' columns in \cref{tab:result2_go,tab:result2_chess}.
Similar to \cref{fig:confusion matrix group n1,fig:confusion matrix group n5,fig:confusion matrix group n10,fig:confusion matrix group n15,fig:confusion matrix group n20}, most predictions were concentrated around $x=y$, suggesting that both method performed well.
However, compared to the random sampling scenario, errors under player-specific estimation tended to spread more across adjacent rank groups.
This indicates a higher variance in individual predictions.

\clearpage

\section{Ablation Studies under Player-Specific Sampling}
\label{appendix:ablation2}

\cref{tab:ablation2_go,tab:ablation2_chess} show the results of the ablation studies under the player-specific sampling scenario that aim for investigating how accuracy under the player-specific sampling scenario was affected by each set of features from strength scores, priors from imitation models, and losses.
The discussions on the results have been made in \cref{subsec:discussions by player}.

{
\tabcolsep = 10pt
\begin{table}[bt]
    \centering
    \caption{Ablation study of our rank estimation meta-model in Go under the player-specific sampling scenario}
    \label{tab:ablation2_go}
      \begin{tabular}{c|cccc}
        \toprule
        $n$ & Use All & w/o Strength & w/o Prior & w/o Loss\\
        \midrule
        5 & \textbf{58.1} (\textbf{95.3}) & 53.1 (93.0) & 52.5 (93.6) & 56.7 (95.1)\\
        10 & 61.0 (96.2) & 56.5 (94.5) & 59.0 (\textbf{96.3}) & \textbf{61.7} (\textbf{96.3})\\
        15 & \textbf{63.2} (96.6) & 57.8 (95.1) & 61.1 (\textbf{97.1}) & 62.8 (96.8)\\
        \bottomrule
      \end{tabular}
\end{table}
}

{
\tabcolsep = 10pt
\begin{table}[bt]
    \centering
    \caption{Ablation study of our rank estimation meta-model in chess under the player-specific sampling scenario}
    \label{tab:ablation2_chess}
      \begin{tabular}{c|cccc}
        \toprule
        $n$ & Use All & w/o Strength & w/o Prior & w/o Loss\\
        \midrule
       5 & \textbf{52.2} (94.5) & 48.8 (92.3) & 47.8 (92.1) & 50.6 (\textbf{94.8})\\
       10 & \textbf{57.0} (\textbf{97.2}) & 54.9 (96.4) & 53.7 (95.1) & 56.6 (96.9)\\
        15 & \textbf{60.4} (\textbf{97.7}) & 60.2 (97.2) & 56.2 (96.0) & 59.8 (97.4)\\
        \bottomrule
      \end{tabular}
\end{table}
}

\section{Box Plots of Prior Geometric Means and Losses}
\label{appendix:boxplot}

\begin{figure}[bt]
    \centering
    \begin{subfigure}[t]{0.49\linewidth}
        \centering
        \raisebox{2mm}{\includegraphics[width=55mm]{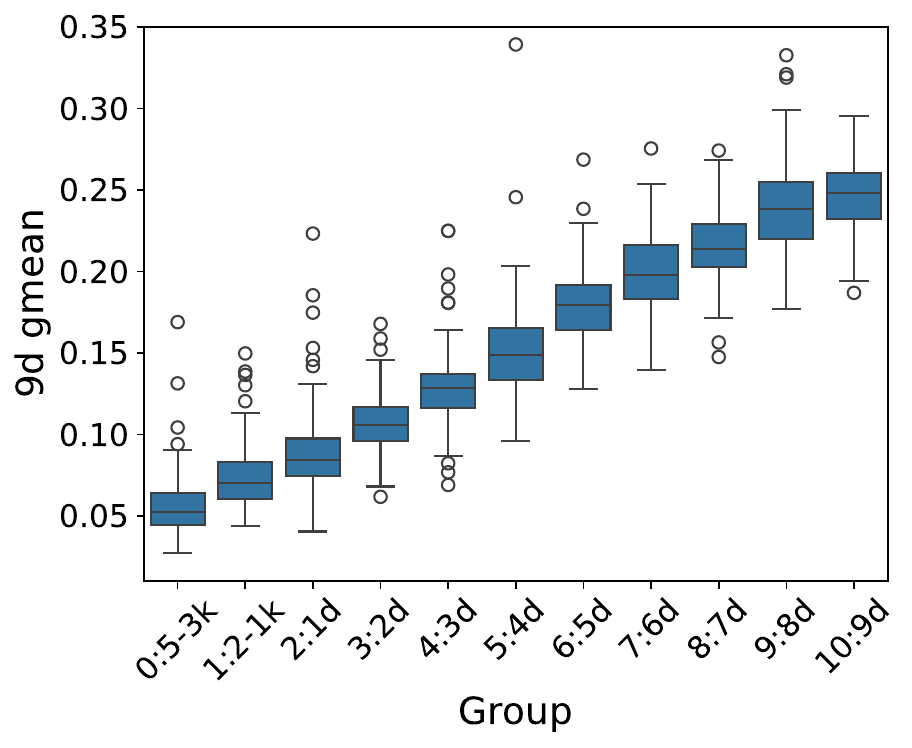}}       
        \caption{Go}
        \label{fig:boxplot_prior_gmean_go}
    \end{subfigure}
    \begin{subfigure}[t]{0.49\linewidth}
        \centering
        \includegraphics[width=55mm]{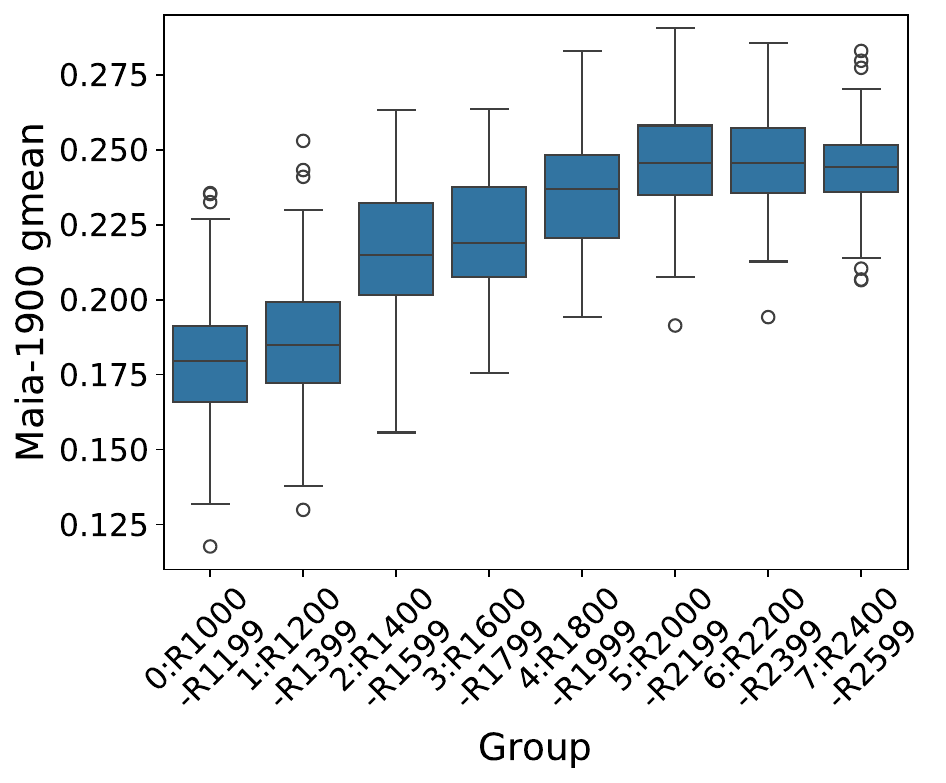}
        \caption{Chess}
        \label{fig:boxplot_prior_gmean_chess}
    \end{subfigure}
    \caption{The prior geometric mean of each player in each rank group with $n=20$, where priors were from KataGo HumanSL with the 9d setting in Go and from Maia-1900 in chess.}
    \label{fig:boxplot_prior_gmean}
\end{figure}

\begin{figure}[bt]
   \centering
    \begin{subfigure}[t]{0.49\linewidth}
        \centering
        \raisebox{2mm}{\includegraphics[width=55mm]{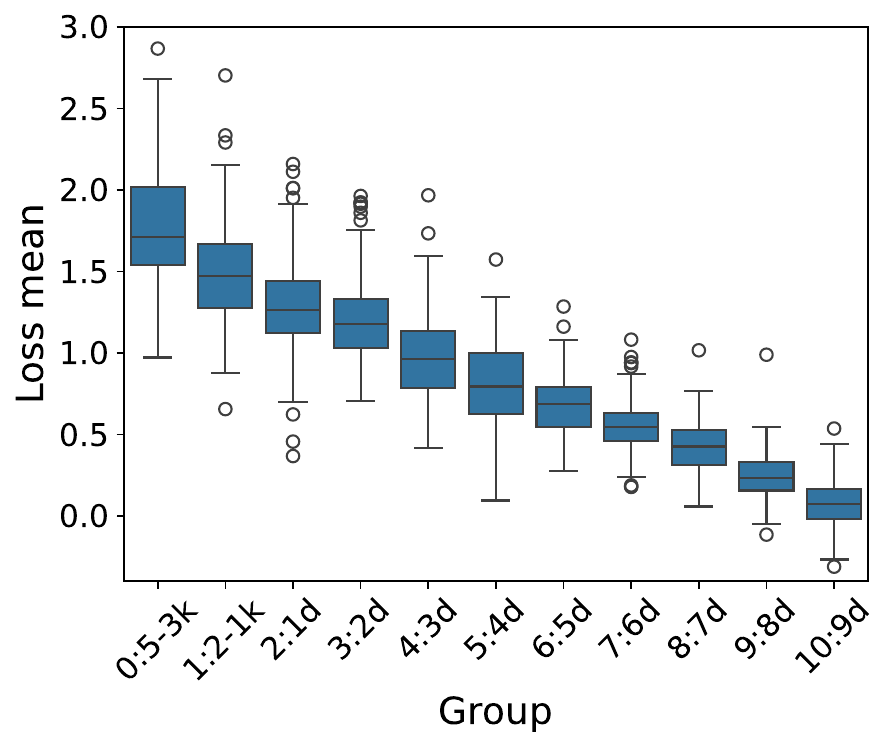}}
        \caption{Go}
        \label{fig:boxplot_loss_mean_go}
    \end{subfigure}
    \begin{subfigure}[t]{0.49\linewidth}
        \centering
        \includegraphics[width=55mm]{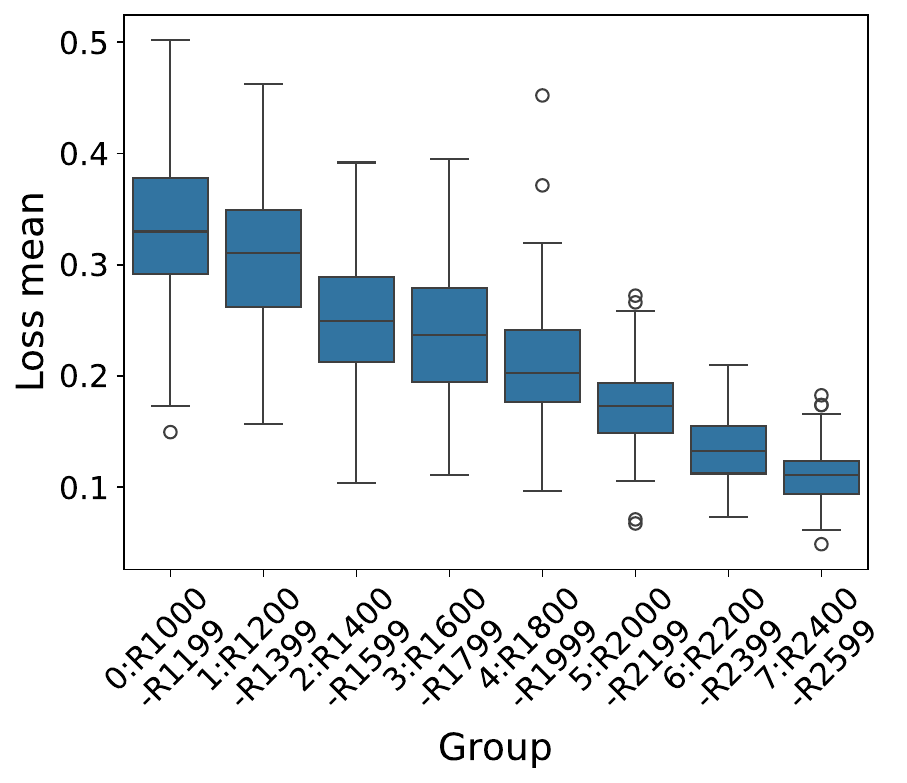}
        \caption{Chess}
        \label{fig:boxplot_loss_mean_chess}
    \end{subfigure}
    \caption{The loss of each player in each rank group with $n=20$.}
    \label{fig:boxplot_loss_mean}
\end{figure}

Similar to \cref{fig:boxplot_str_gmean} for mean strength scores, we depicted box plots for prior geometric means and losses.
In more detail, we collected the prior geometric mean and the loss from 20 sampled data points for each player and repeated this process for all 100 players in every rank group.
For prior geometric means, we used KataGo HumanSL with the 9d setting in Go and from Maia-1900 in chess.
\cref{fig:boxplot_prior_gmean,fig:boxplot_loss_mean} show the box plots for prior geometric means and losses, respectively.

Both prior geometric means and losses showed greater overlap between rank groups compared to strength scores.
The results indicated that both prior geometric means from a single model and losses were less powerful to distinguish players in adjacent rank groups when used individually.
Nevertheless, as observed in \cref{subsec:discussions main}, combining prior geometric means from multiple imitation models significantly improved rank estimation accuracy.
This suggests that rank estimation accuracy may be improved by incorporating additional features, even those with limited individual effectiveness.

\end{document}